\pdfoutput=1

\documentclass[11pt]{article}

\usepackage[final]{acl}

\usepackage{times}
\usepackage{latexsym}
\usepackage[normalem]{ulem}
\useunder{\uline}{\ul}{}
\usepackage[T1]{fontenc}

\usepackage[utf8]{inputenc}

\usepackage{microtype}

\usepackage{inconsolata}

\usepackage{bm}
\usepackage{tikzducks}
\usepackage{graphicx}
\usepackage[most]{tcolorbox}
\usepackage{colortbl}
\usepackage{pifont}
\usepackage{enumitem}
\definecolor{Gray}{gray}{0.9}

\usepackage{listings}
\definecolor{codeblue}{rgb}{0.25,0.5,0.5}
\definecolor{codekw}{rgb}{0.85, 0.18, 0.50}
\lstdefinestyle{mystyle}{
    backgroundcolor=\color{white},
    basicstyle=\fontsize{7.5pt}{7.5pt}\ttfamily\selectfont,
    columns=fullflexible,
    breaklines=true,
    captionpos=b,
    commentstyle=\fontsize{7.5pt}{7.5pt}\color{codeblue},
    keywordstyle=\fontsize{7.5pt}{7.5pt}\color{codekw},
}
\usepackage{graphicx}
\usepackage{amsmath, amssymb, bm}
\usepackage{algorithm}
\usepackage{multirow}
\usepackage{graphicx}
\usepackage{booktabs}
\usepackage{multicol}
\usepackage{hyperref}
\usepackage{amsmath}
\usepackage{caption}
\usepackage{enumitem}
\usepackage{wrapfig}
\usepackage{listings}
\usepackage{pifont}
\usepackage{array}
\usepackage{placeins}
\usepackage{float}
\definecolor{codegreen}{rgb}{0.0, 0.411, 0.243}
\definecolor{codered}{rgb}{0.89, 0.26, 0.20}
\usepackage{hyperref}
\definecolor{dartgreen}{HTML}{00693e}

\definecolor{refcolor}{HTML}{B3A369}
\hypersetup{
    colorlinks=true,
    linkcolor=refcolor,
    citecolor=refcolor,
    filecolor=magenta,      
    urlcolor=refcolor,
    }

\title{Superficial Self-Improved Reasoners Benefit from Model Merging}

\author{
  Xiangchi Yuan$^{1}$,
  Chunhui Zhang$^{2}$,
  Zheyuan Liu$^{3}$, 
  Dachuan Shi$^{1}$, \\
  \textbf{Leyan Pan$^{1}$,
  Soroush Vosoughi$^{2}$,
  Wenke Lee$^{1}$} \\
  $^{1}$Georgia Institute of Technology,
  $^{2}$Dartmouth College,
  $^{3}$University of Notre Dame \\
  \texttt{\{xyuan300, dshi77, leyanpan\}@gatech.edu, wenke@cc.gatech.edu} \\
  \texttt{\{chunhui.zhang.gr, soroush\}@dartmouth.edu} \\
  \texttt{zliu29@nd.edu}
}

\begin{document}
\maketitle

\begin{abstract}
As scaled language models (LMs) approach human-level reasoning capabilities, self-improvement emerges as a solution to synthesizing high-quality data corpus. While previous research has identified model collapse as a risk in self-improvement, where model outputs become increasingly deterministic, we discover a more fundamental challenge: \textit{the superficial self-improved reasoners phenomenon}. In particular, our analysis reveals that even when LMs show improved in-domain (ID) reasoning accuracy, they actually compromise their generalized reasoning capabilities on out-of-domain (OOD) tasks due to \textit{memorization} rather than genuine \textit{learning}.
Through a systematic investigation of LM architecture, we discover that during self-improvement, LM weight updates are concentrated in less reasoning-critical layers, leading to superficial learning. To address this, we propose Iterative Model Merging (IMM), a method that strategically combines weights from original and self-improved models to preserve generalization while incorporating genuine reasoning improvements. Our approach effectively mitigates both LM collapse and superficial learning, moving towards more stable self-improving systems. Code is available\footnote{Code is available at \href{https://github.com/xiangchi-yuan/merge_syn}{IMM}.}.
\end{abstract}

\section{Introduction}
\label{sec:intro}

The reasoning capabilities~\cite{jaech2024openai, zhang2024working, guo2025deepseek} of large language models (LLMs) largely benefits from vast amounts of high-quality reasoning data. However, as the data corpus runs out \cite{ilya2024sequence} and increasingly powerful models approach human-level intelligence \cite{deepmind2023imo,deepmind2023alphazero}, pressing issues emerge: \textit{(i)} How to advance models' reasoning capabilities despite data scarcity? \textit{(ii)} How to obtain training data that exceeds human-level performance for next-generation models? A promising answer to both questions is model self-improvement or self-evolution, where models autonomously generate infinite high-quality data, which potentially surpasses human annotations, to continuously enhance their own performance.

\begin{figure}[t]
    \centering
    \includegraphics[width=0.5\textwidth]{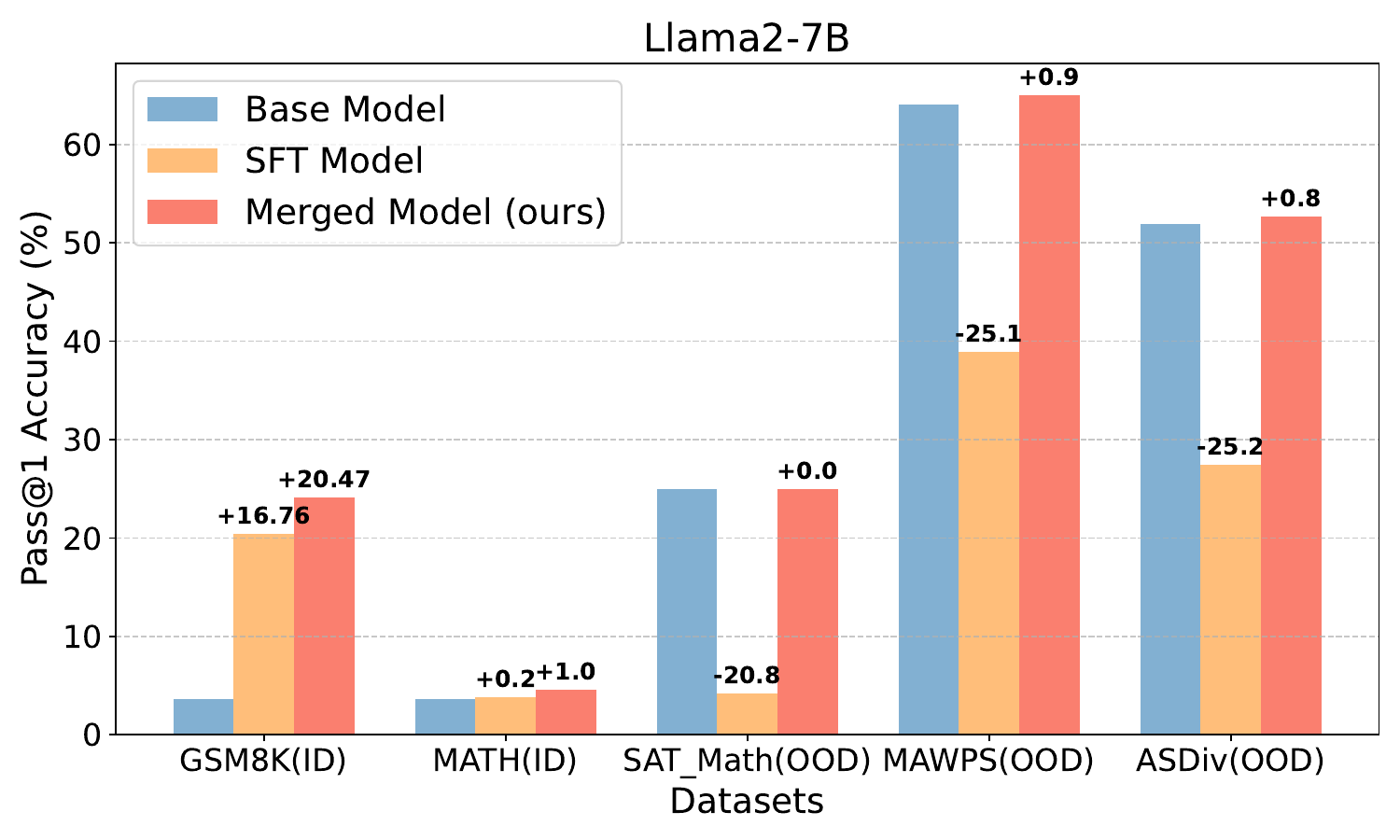} 
    \caption{The Superficial Self-Improved Reasoners phenomenon is mitigated by iterative model merging. Our method improves ID and OOD reasoning performances.}
    \label{fig:intro}
    \vspace{-5pt}
\end{figure}

Although self-improvement has achieved remarkable success in specific domains such as mathematics \cite{openai2025o3, deepmind2023imo}, coding \cite{li2022competition}, and games \cite{hu2024gamearena, silver2018general}, recent studies reveal significant risks associated with using self-generated synthetic data for fine-tuning: in particular, model performance can degrade over multiple iterations of self-improvement, a phenomenon known as \textit{model collapse}. \cite{shumailov2023curse}. In current research, model collapse is primarily attributed to a reduction in sampling diversity \cite{shumailov2023curse, alemohammad2024selfconsuming, guo-etal-2024-curious}. To mitigate this problem, several studies suggest refreshing synthetic data with real data \cite{bertrand2024on, alemohammad2024selfconsuming}, accumulating data across training steps \cite{gerstgrasser2024is}, and incorporating data verifiers \cite{gillman2024selfcorrecting} or correctors \cite{feng2025beyond}.
However, by focusing solely on data quality and diversity, these approaches overlook a more critical question: whether self-improvement genuinely enhances reasoning capabilities or merely memorizes the training distribution. This distinction becomes crucial when considering the model's ability to generalize beyond its training data.

In this paper, we investigate a risk in model self-improvement for reasoning tasks that deepens the known challenge of model collapse. We identify a phenomenon we call Superficial Self-Improved Reasoners, where models appear to improve but actually fail to develop genuine reasoning capabilities. While these models show enhanced performance on in-domain (ID) reasoning tasks, they significantly underperform on out-of-domain (OOD) tasks, suggesting memorization rather than genuine reasoning improvement.
To understand the mechanistic cause of this phenomenon, we perform a systematic analysis of the model architecture during self-improvement. By examining layer importance and parameter changes, we uncover a critical mismatch: the largest weight updates occur in layers that contribute least to reasoning, while reasoning-critical layers receive minimal updates. This mismatch explains why models tend to memorize training patterns rather than develop generalizable reasoning skills.
To address this issue, we propose Iterative Model Merging (IMM), a novel method that strategically combines weights from original and self-improved models. IMM specifically targets the layer misalignment problem by preserving the stability of reasoning-critical layers while allowing beneficial updates from self-improvement. As demonstrated in Figure~\ref{fig:intro}, this approach effectively balances performance improvements with preserved generalized reasoning capability. 

A summary of the contributions is given below:

\begin{itemize}[leftmargin=*] \item This work identifies the risk of self-improvement for reasoning: while the model enhances its reasoning capabilities, it still tends to memorize the training data, resulting in a loss of generalized reasoning ability. We refer to this phenomenon as \textit{Superficial Self-Improved Reasoners}. \item We provide an explanation for this phenomenon by highlighting a mismatch between the reasoning-critical layers and the layers that undergo the largest weight changes.  \item We propose IMM to mitigate this phenomenon. IMM offers a simple, general, and effective approach to integrate the reasoning improvements of the self-improved model while preserving the generalization of the original model. \end{itemize}

\begin{figure*}[h!]
    \vspace{12pt}
    \centering
    \begin{minipage}[t]{0.32\textwidth}
        \centering
        \includegraphics[width=\linewidth]{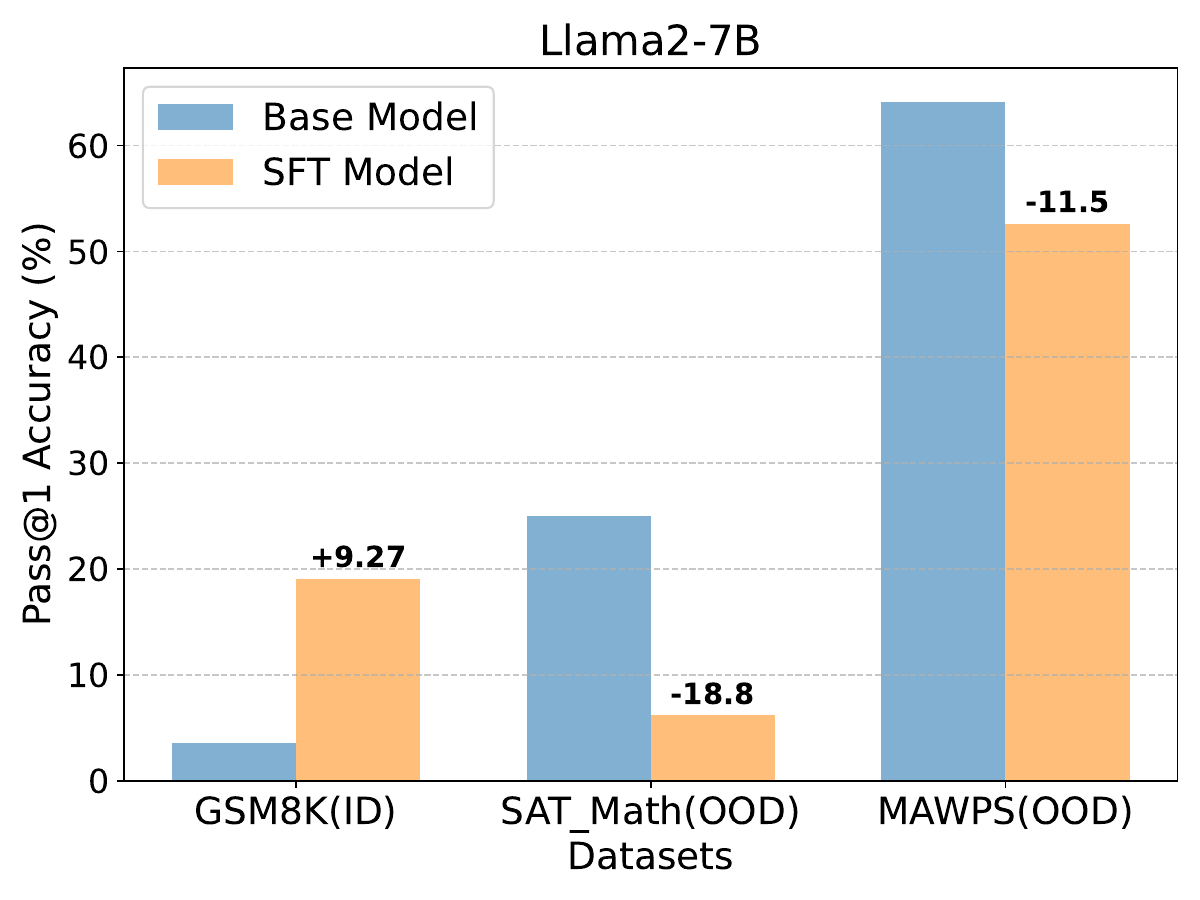}
    \end{minipage}%
    \begin{minipage}[t]{0.32\textwidth}
        \centering
        \includegraphics[width=\linewidth]{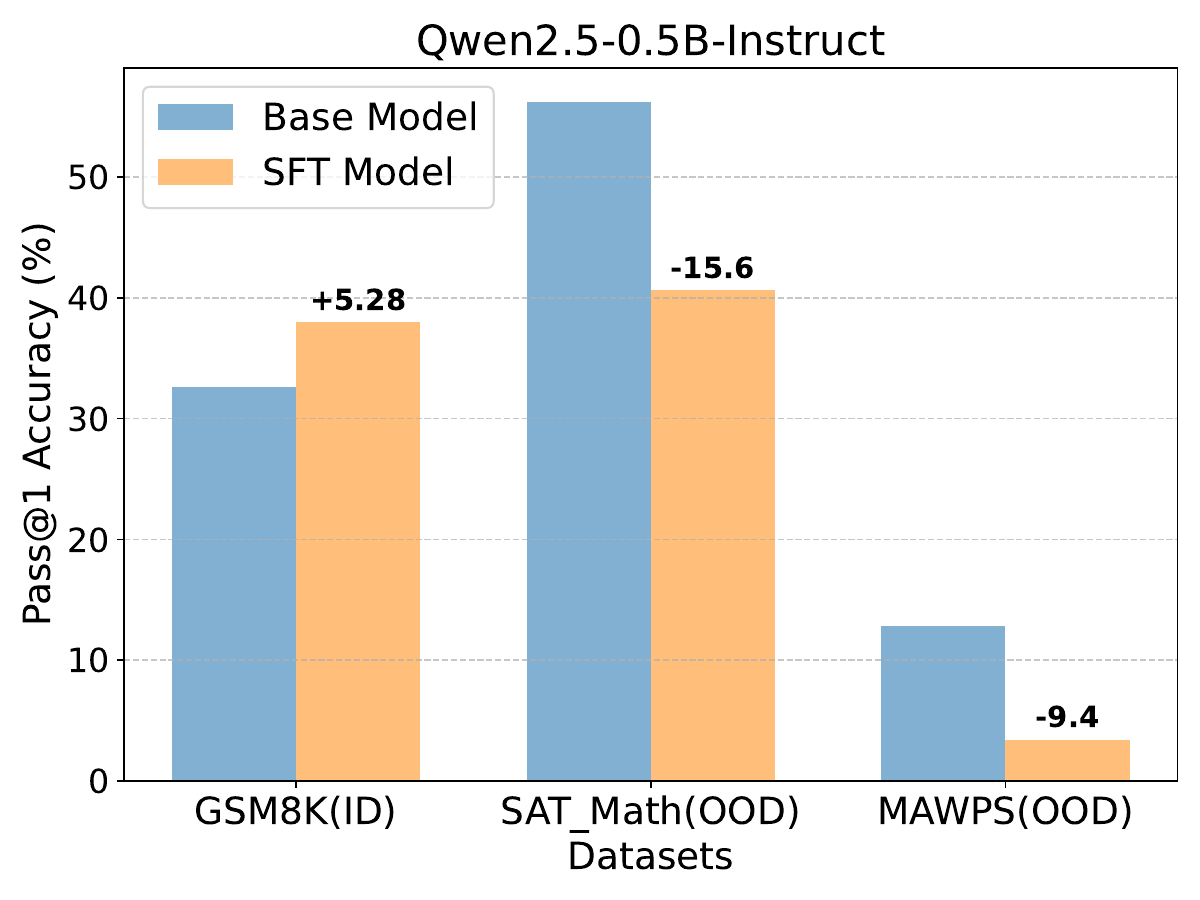}
    \end{minipage}%
    \begin{minipage}[t]{0.32\textwidth}
        \centering
        \includegraphics[width=\linewidth]{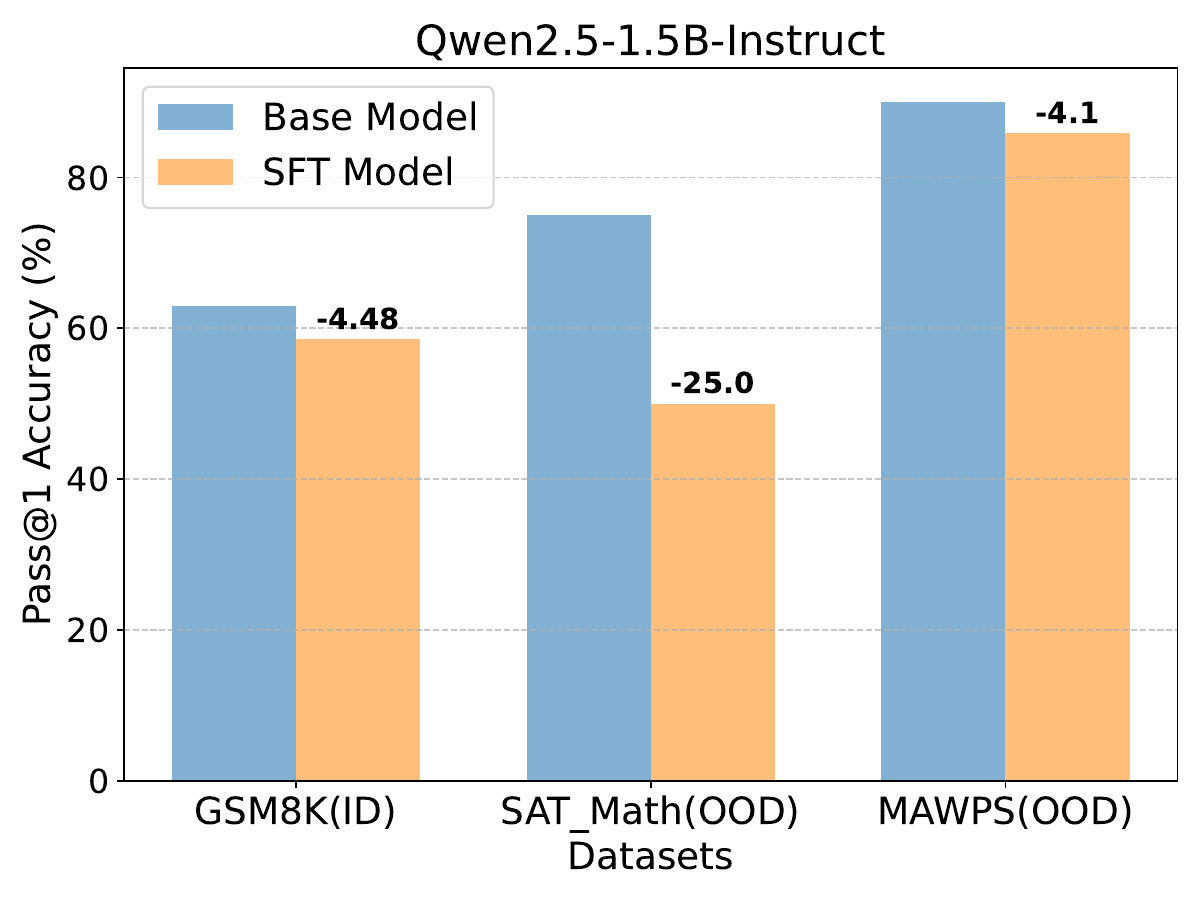}
    \end{minipage}%
    \hspace{-1em}
 
    \caption{Superficial Self-improved Reasoners. The model's performance is only improved on in-domain reasoning datasets while losing the generalized reasoning capabilities on out-of-domain reasoning datasets.}
    
    \label{fig:id_ood}

\end{figure*}

\section{Related Work}
\label{sec:related}



\paragraph{LLM Self-Improvement}
Given the high cost of labeling data, it is increasingly common to leverage LLMs to generate synthetic responses for training student models. Traditionally, this process has focused on knowledge distillation from stronger teacher models \cite{yuan2023scaling, wu2024meta}. More recently, studies have demonstrated that distilling from weaker models—referred to as weak-to-strong knowledge distillation—can be more beneficial for LLMs compared to distilling from stronger models, given the same computational budget \cite{bansal2024smaller}.
Another emerging direction is LLM self-improvement, where models improve themselves using their own outputs \cite{huang2022large, gulcehre2023reinforced, singh2023beyond}. In the context of reasoning tasks, various self-improvement methods have been proposed: SPO \cite{prasad2024self} employs Self-Consistency Preference Optimization for self-improvement; \citet{pang2024iterative} iteratively generate and refine data to optimize the model’s reasoning ability; and \citet{hosseini2024v} utilize both correct and incorrect answers to improve reasoning performance through training an additional verifier.


\paragraph{Model Collapse}
As real-world data becomes increasingly scarce \cite{ilya2024sequence}, synthetic data is playing a crucial role in training modern generative models due to its low cost and infinite availability. However, recent studies have revealed the risks associated with this "free lunch," a phenomenon known as model collapse \cite{shumailov2023curse}. The model collapse has been extensively identified and analyzed in both computer vision \cite{hataya2023will, he2022synthetic, bohacek2023nepotistically} and natural language processing \cite{alemohammad2024selfconsuming, gerstgrasser2024is}. Researchers have investigated its underlying causes from both empirical \cite{padmakumar2024does, guo2023curious} and theoretical perspectives \cite{yuan2024self, bertrand2023stability, seddik2024bad,fu2024towards}.
Current approaches to mitigating model collapse predominantly focus on data-centric methods. \citet{feng2025beyond} show that imperfect verifiers can help prevent model collapse by selecting appropriate data. \citet{shumailov2023curse} proposes mixing data from previous iterations to prevent performance degradation, while \citet{gerstgrasser2024is} demonstrates that accumulating synthetic data over iterations reduces the risk of collapse.

Appendix~\ref{app:realted_works} discusses additional related works on LLM for reasoning. The connection with catastrophic forgetting is discussed in Appendix~\ref{app:catastrophic}.
\section{Superficial Self-improved Reasoners}

\label{sec:motivation}


A natural and critical question arises for LLM self-improvement: does learning from synthetic reasoning data generated by the model itself trade off generalization ability for improved reasoning performance because of learning from itself? Our study shows that the answer is yes.
In this section, we first confirm that self-improvement enhances in-domain reasoning performance but degrades general reasoning capabilities. We then investigate the underlying cause of this phenomenon by analyzing the layer-wise importance of the model during reasoning and tracking weight changes throughout the self-improvement process. A detailed comparison reveals a notable mismatch: the layers most crucial for reasoning experience relatively small weight updates, while less critical layers undergo more significant changes. This suggests that strong reasoning layers fail to substantially improve their reasoning ability through weight updates, whereas less important layers tend to overfit the training data rather than truly learning to reason.

\subsection{Identify Superficial Self-improved Reasoners from OOD datasets}
In this part, we identify  Superficial Self-improved Reasoners by self-improving LLMs on the ID reasoning datasets and test them on OOD datasets.
\paragraph{Synthesizing Reasoning Data for Self-improvement}

We begin by establishing the self-improvement framework through the generation of reasoning data. Following prior work \cite{zelikman2022star}, we first synthesize reasoning data for fine-tuning. Let \( \mathcal{D} = \{(q_i, a_i)\}_{i=1}^{n_d} \) denote a training dataset containing \( n_d \) reasoning questions \( q_i \) and corresponding final answers \( a_i \). We also use Chain-of-Thought prompting \cite{wei2022chain} in this process (details in Appendix~\ref{app:cot}). In the second step, we sample multiple solutions for each \( q_i \) using non-zero sampling temperatures, resulting in a synthetic dataset \( \mathcal{D}_S = \{(q_i, \{(\hat{r}_{ij}, \hat{a}_{ij})\}_{j=1}^{k})\} \), where \( k \) represents the number of sampled solutions. Here, \( \hat{r}_{ij} \) denotes the \( j \)-th reasoning path (i.e., rationale) generated by the model for \( q_i \), and \( \hat{a}_{ij} \) is the model’s corresponding final answer.  Incorrect solutions are then filtered out by comparing the sampled answers \( \hat{a}_{ij} \) with the ground-truth answers \( a_i \). Finally, we fine-tune the model on the filtered dataset \( \tilde{\mathcal{D}}_G \) using supervised fine-tuning (SFT) to maximize the likelihood of generating reasoning paths \( r \), optimizing the following objective:
\begin{equation}
    \mathbb{E}_{(q,r,a) \sim \tilde{\mathcal{D}}_G} \left[ \log p_{\theta}(r, a | q) \right].
\end{equation}


\paragraph{Loss of Generalized Reasoning Ability during Self-Improvement}
After applying the self-improvement framework to LLMs of various scales on ID datasets, we evaluate their performance on OOD reasoning datasets. The results, presented in Figure \ref{fig:id_ood}, reveal that while self-improvement enhances reasoning performance on ID datasets, it leads to a noticeable decline in performance on OOD datasets. This phenomenon suggests that although self-improvement improves metrics on ID reasoning tasks, it fails to enhance generalized reasoning capabilities and may even degrade them. We refer to this behavior as the emergence of \textit{Superficial Self-Improved Reasoners}.



\subsection{Investigating the Causes of Superficial Self-Improved Reasoners}

\begin{figure}[t]
    \centering
    \begin{minipage}[t]{0.25\textwidth}
        \centering
        \includegraphics[width=\linewidth]{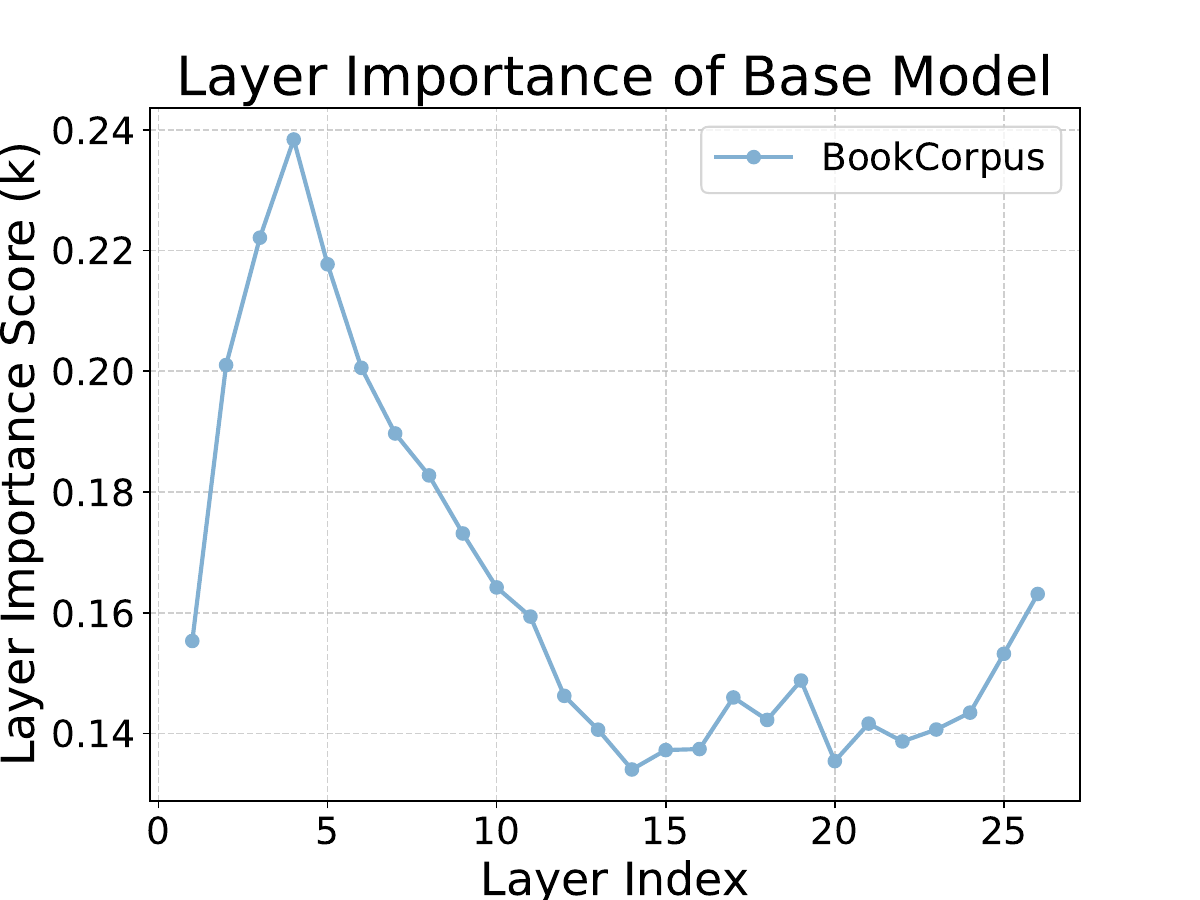}
    \end{minipage}%
    \begin{minipage}[t]{0.25\textwidth}
        \centering
        \includegraphics[width=\linewidth]{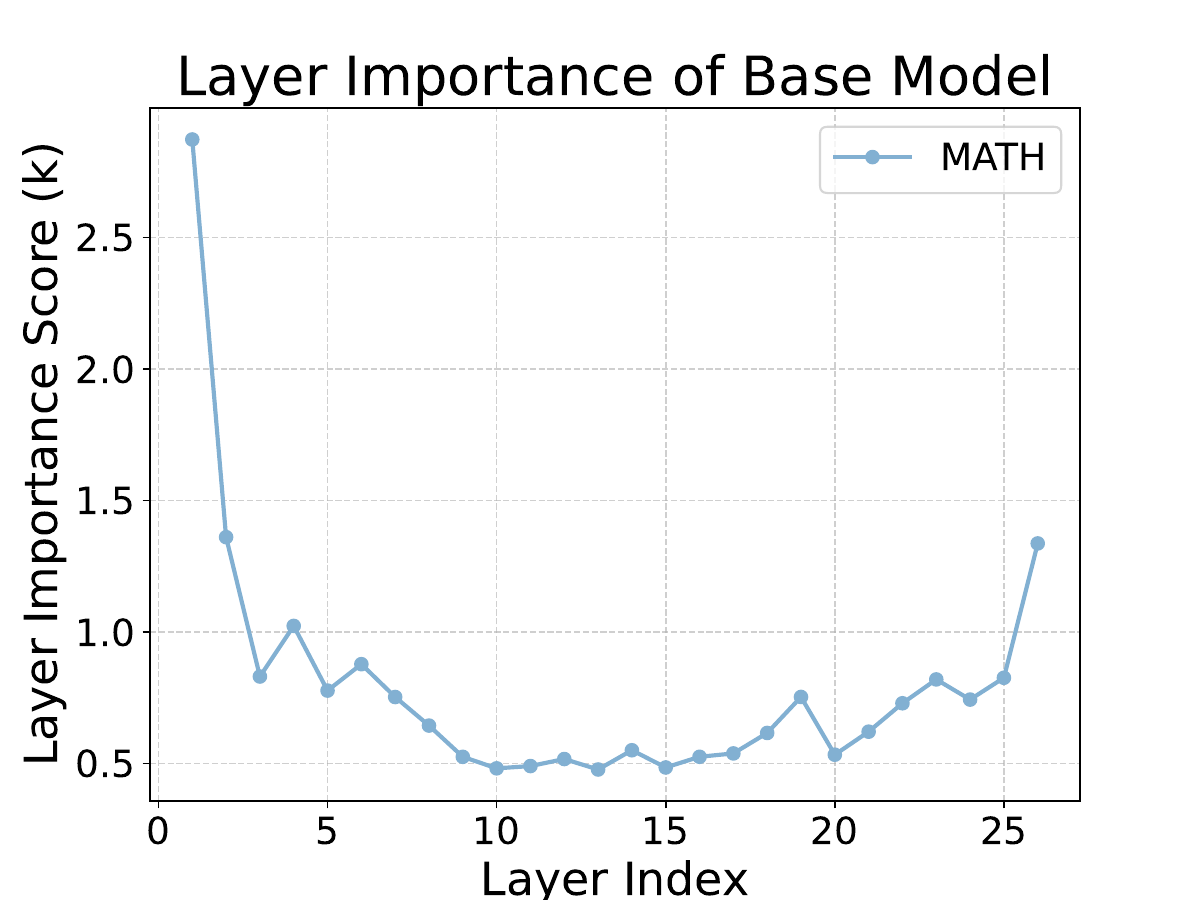}
    \end{minipage}%
    
    \caption{The Layer Importance Scores of strong reasoning model Qwen2.5-1.5B-Math on BookCorpus (left) and MATH datasets (right). The middle layers are less important while the early and late layers are more important for reasoning (MATH). For non-reasoning task (BookCorpus) middle layers are more important.}
    \label{layer_importance}
    
\end{figure}

While numerous studies on catastrophic forgetting focus on analyzing and addressing OOD performance degradation in continual learning for learning simpler tasks, our work specifically targets the more challenging domain of mathematical reasoning in LLMs, with an emphasis on understanding the phenomenon of Superficial Self-Improved Reasoners. In this section, we identify the most critical layers for reasoning, analyze how their weights evolve during the self-improvement process, and provide an explanation for the emergence of Superficial Self-Improved Reasoners.




\paragraph{Layer Importance for Reasoning}  
To identify the most important weights in LLMs for reasoning, our objective is to determine and remove the weights that have the greatest impact on the model’s prediction, which can be measured by the resulting change in loss. We denote the linear weight matrix as $\mathbf{W}^{k,n} =
\begin{bmatrix}
W^{k,n}_{i,j}
\end{bmatrix},$
where \( k \) represents the modules (e.g., a key projection in the multi-head attention (MHA) or an up-projection in the feed-forward network (FFN)) within the \( n \)-th LLM layer.  
We quantify the importance of each weight by measuring the error introduced when the corresponding parameter is removed. Given an in-domain reasoning dataset \( \mathcal{D} \), the importance score \( I^{k,n}_{i,j} \) for the weight \( W^{k,n}_{i,j} \) is defined as:  
\begin{equation}
    \begin{aligned}
I^{k,n}_{i,j} &= \left | \Delta \mathcal{L} (\mathcal{D}) \right | \\  &= \left| \frac{\partial \mathcal{L} (\mathcal{D})}{\partial W_{i,j}^{k,n}} W_{i,j}^{k,n} 
- \frac{1}{2} W_{i,j}^{k,n} H_{kk} W_{i,j}^{k,n} \right. \\
&\quad \left. + \mathcal{O} \left( \|W_{i,j}^{k,n}\|^3 \right) \right|.
\end{aligned}
\end{equation}

\begin{figure}[t]
    \centering
    \begin{minipage}[t]{0.25\textwidth}
        \centering
        \includegraphics[width=\linewidth]{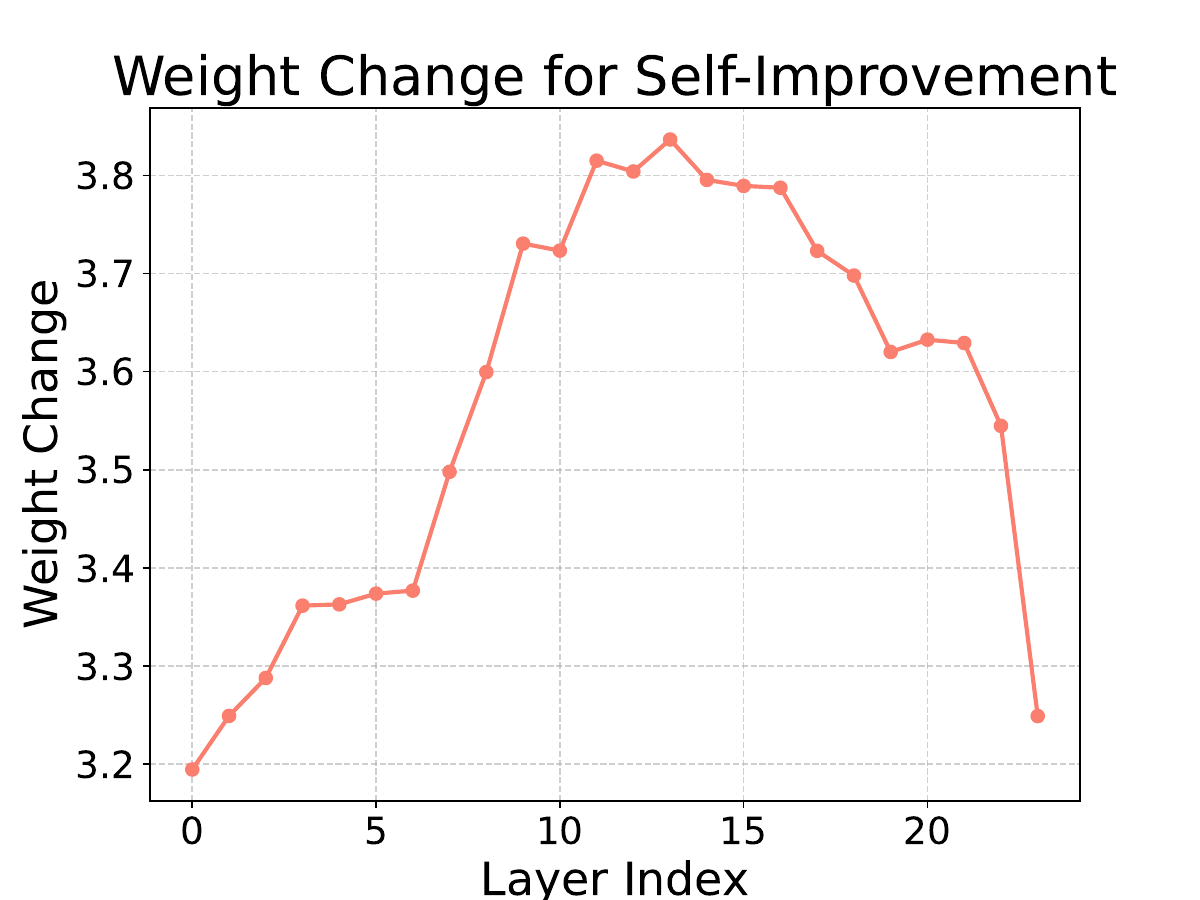}
    \end{minipage}%
    \begin{minipage}[t]{0.25\textwidth}
        \centering
        \includegraphics[width=\linewidth]{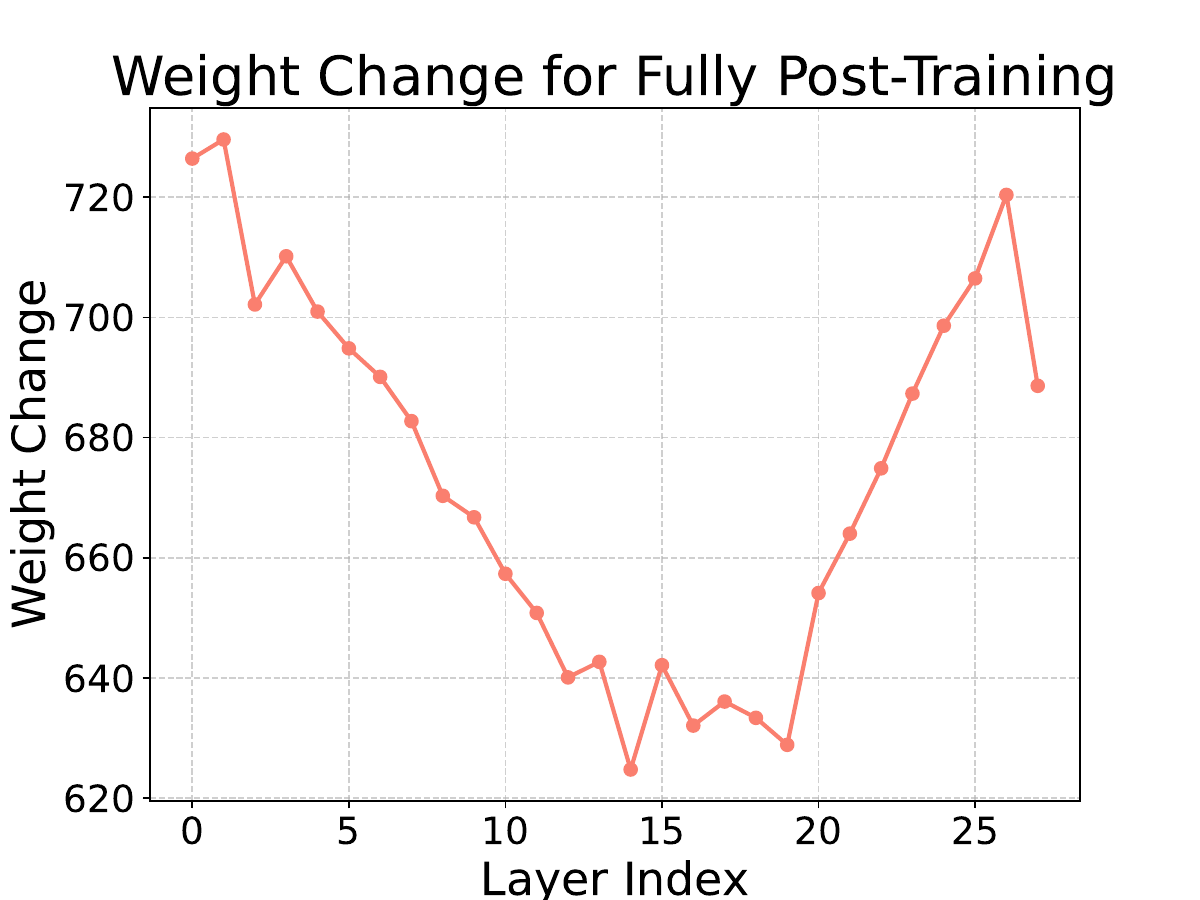}
    \end{minipage}%
    
    \caption{The weight change for SFT Qwen2.5-1.5B with self-improvement MATH data (left) and fully post-training Qwen2.5-1.5B to Qwen2.5-1.5B-Math using real data with 700B tokens (right).}
    \label{weight_change}
   
\end{figure}
However, due to the significant computational cost associated with the large number of parameters in LLMs, we approximate the Hessian matrix \( H_{kk} \) using the Fisher information matrix, following the approach in \citet{ma2023llm}. This allows us to approximate the second-order term \( \frac{1}{2} W_{i,j}^{k,n} H_{kk} W_{i,j}^{k,n} \) as \( \frac{1}{2} \sum_{j=1}^{N} \left( \frac{\partial \mathcal{L} (\mathcal{D}_j)}{\partial W_{i}^{k}} W_{i}^{k} \right)^2 \). By omitting the second-order derivative, the importance score \( I^{k,n}_{i,j} \) is simplified to: $I^{k,n}_{i,j} \approx \left| \frac{\partial \mathcal{L} (\mathcal{D})}{\partial W^{k,n}_{i,j}} W^{k,n}_{i,j} \right|$. To assess the contribution of each layer to reasoning, we define the layer importance score as:  
\begin{equation}
\label{eq:importance}
    I^n = \sum_{W^{k,n}_{i,j}} \left| \frac{\partial \mathcal{L}(\mathcal{D})}{\partial W^{k,n}_{i,j}} W^{k,n}_{i,j} \right|.
\end{equation}
We leverage this layer importance score $I^n$ to identify which layers contribute most significantly to reasoning tasks. As illustrated in Figure \ref{layer_importance}, the middle layers are less
important while the early and late layers are more important for the reasoning (MATH) tasks. We also find similar performance on code reasoning tasks, as illustrated in Appendix~\ref{app:exp_layer_i}. However, for the non-reasoning dataset BookCorpus, the middle layers are more important. This observation highlights the early and late layers as \textit{reasoning-critical layers} (More clarification for this term is in Appendix~\ref{app:def_layers}), distinguishing their specialized function in reasoning.


\paragraph{Layer Weight Change after Self-Improvement}  
After fine-tuning the LLMs on reasoning data, the weights are updated, enabling the model to learn reasoning capabilities. We now analyze these weight changes. Let \( \Delta \mathbf{W}^{n} \) represent the total weight change at the \( n \)-th layer after SFT:  
\begin{equation}
\label{eq:weight_change}
    \Delta \mathbf{W}^{n}  =  \sum_{k}  \left \| \mathbf{W}^{k,n} - \mathbf{W}^{k,n}_{\text{SFT}}  \right \|,
\end{equation}
where \( \mathbf{W}^{k,n} \) denotes the original $k$-th weight matrix and \( \mathbf{W}^{k,n}_{\text{SFT}} \) is the fine-tuned weight matrix. Figure \ref{weight_change} illustrates the weight change \( \Delta \mathbf{W}^{n} \) across different layers. For the self-improved model, the largest weight change occurs in the middle layers. In contrast, for the math model which is fully post-trained with stronger generalized reasoning capability, the most significant weight changes are concentrated in the early and late layers. 
A similar condition happens for real data with limited training data size, as analyzed in Appendix~\ref{app:real_data}.

\begin{figure*}[t]
    \centering
    \includegraphics[width=1\textwidth]{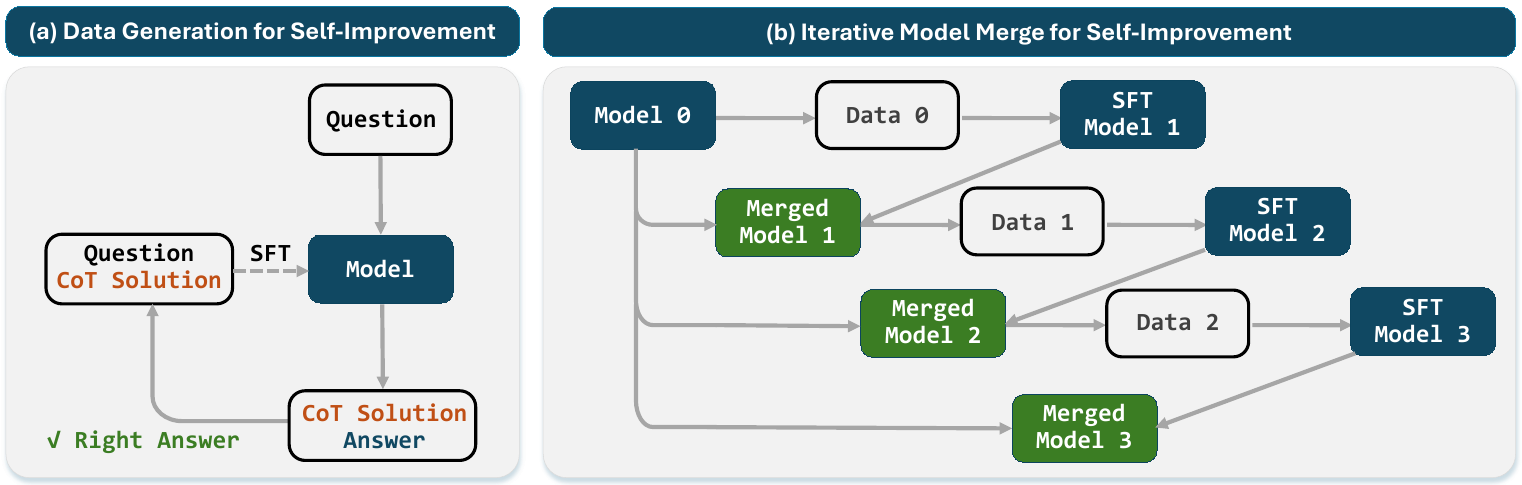} 
    \caption{The overall framework: \textbf{(a)} The model generates chain-of-thought (CoT) answers for the given questions, and incorrect answers are filtered out using the ground-truth. The remaining correct answers are used for SFT to self-improve the model. \textbf{(b)} IMM iteratively SFT the model and merges the self-improved models with the base model to balance reasoning enhancement and generalization.}
    \label{fig:method}
\end{figure*}

\begin{table}[]
\centering
\resizebox{0.5\textwidth}{!}{%
\begin{tabular}{lccc}
\toprule
\textbf{Model}     & \textbf{\begin{tabular}[c]{@{}c@{}} Reasoning-\\Critical Layer\end{tabular}} & \textbf{\begin{tabular}[c]{@{}c@{}}Most Weight \\ Change Layer\end{tabular}} & \textbf{\begin{tabular}[c]{@{}c@{}}Generalized \\ Reasoning Capability\end{tabular}} \\ \hline
Self-Improved      & Early,  late                                                   & Middle                                                       &   \ding{55}                                           \\
Fully Post-trained & Early,  late                                                  & Early,  late                                                 &      $\checkmark$                                               \\ \bottomrule
\end{tabular}
}
\caption{Comparison of self-improved model and fully post-trained math model.}
\vspace{-5pt}
\label{tab:comparison}
\end{table}


\paragraph{Takeaway}
By analyzing Figure \ref{layer_importance} and Figure \ref{weight_change} (left), we observe that the middle layers (reasoning-trivial layers) are the least important for the strong reasoning capabilities of LLMs, yet these layers undergo the most significant updates during the self-improvement process. This phenomenon highlights a contradiction in how reasoning ability is acquired. If the model were solely learning generalized reasoning, the most substantial weight updates would occur in the early and late layers (reasoning-critical layers), as observed in fully post-trained math models with strong generalized reasoning capabilities, rather than in the middle layers.

\noindent This observation suggests that during self-improvement, the model does not exclusively enhance its reasoning ability but also exhibits a tendency to overfit the training data, effectively "memorizing" it. This overfitting behavior explains the improved performance on ID datasets while compromising the model’s generalization to OOD tasks. The performance comparison in Figure \ref{fig:id_ood} further supports this conclusion. We summarize all experimental findings in Table \ref{tab:comparison}, which leads to the following key insights: \textit{(i)} during self-improvement on reasoning tasks, LLMs may show improved reasoning performance on ID tasks but lose generalized reasoning ability on OOD tasks; \textit{(ii)} This phenomenon arises from a mismatch between the reasoning-critical layers and the layers with significant weight changes, suggesting that the model memorizes the training data rather than truly learning generalized reasoning capability. We further provide analysis on the reasons for this\textbf{ mismatch phenomenon} in Appendix~\ref{app:why_mismatch}.

\section{Superficial Self-improved Reasoners Benefit from \textit{Iterative Model Merging}}


\paragraph{Iterative Model Merging (IMM)} 
In this section, we propose Iterative Model Merging (IMM) to mitigate the \textit{Superficial Self-Improved Reasoners} phenomenon, as illustrated in Figure~\ref{fig:method}. In the first self-improvement iteration, we self improve the original base model and merge the resulting SFT model \( \boldsymbol{\theta_{SFT}^{0}} \) with the base model \( \boldsymbol{\theta} \) to obtain the merged model \( \boldsymbol{\theta_{m}^{0}} \).  
In each subsequent iteration \( t \) (\( t > 0 \)), we continue the self-improvement process by fine-tuning the previously merged model \( \boldsymbol{\theta_{m}^{t-1}} \). The resulting self-improved model \( \boldsymbol{\theta_{SFT}^{t}} \) is then merged with the original base model to obtain the updated merged model \( \boldsymbol{\theta_{m}^{t}} \).  
To formally describe this process, we define the parameter change \( \boldsymbol{\delta^t} \) during each SFT iteration as follows:
\begin{equation}
    \boldsymbol{\delta^t} =
    \begin{cases}
  \boldsymbol{\theta^t_{SFT}}-\boldsymbol{\theta_m^{t-1}}, & \text{if } t > 0 \text{, SFT } \boldsymbol{\theta_m^{t-1}}, \\
  \boldsymbol{\theta^t_{SFT}}-\boldsymbol{\theta},  & \text{if } t = 0 \text{, SFT } \boldsymbol{\theta}.
\end{cases}
\end{equation}
We then incorporate DARE \cite{yu2024language} to further process \( \boldsymbol{\delta^t} \). DARE identifies parameter redundancy in LLMs, randomly masking parameter changes at a drop rate \( p \) while scaling the remaining updates to improve the performance of the merged model. Denoting \( \mathbf{m} \sim \text{Bernoulli}(p) \), DARE can be expressed as:  
\[
\tilde{\boldsymbol{\delta^t}} = (1 - \mathbf{m}) \odot \boldsymbol{\delta^t}, \quad 
\hat{\boldsymbol{\delta^t}} = \tilde{\boldsymbol{\delta^t}} / (1 - p).
\]  
By incorporating DARE into our iterative model merging framework, the final update for each iteration \( t \) is given by:  
\begin{equation}
    \boldsymbol{\theta_m^{t+1}} = \alpha \boldsymbol{\theta} + (1-\alpha)(\boldsymbol{\theta^{t}} + \boldsymbol{\hat{\delta^t}}),
\end{equation}
where \( \alpha \) is a scaling parameter that controls the balance between the base model weights and the self-improved model weights. Although we use a uniform \( \alpha \) for all layers, which makes reasoning-critical layers's weight change remain minimal at the first iteration, this generalized way makes the model avoid overfitting and learn the generalized reasoning capability, which makes reasoning-critical layers' weights change increase more compared to the reasoning-trivial layers in the next iterations to learn generalized reasoning capability, as analyzed in Appendix~\ref{app:weight_change_iterations}. The overall merging strategy is scalable for multiple iterations and larger models, with complexity analysis presented in Appendix~\ref{app:complexity}.

\paragraph{Insights for IMM} The rationale behind model merging for generalized reasoning capability can be understood from two perspectives:
\textit{(i)} Based on the experimental observations in Section \ref{sec:motivation}, the weights of reasoning-critical layers undergo significant changes during self-improvement, indicating that these layers are likely memorizing the training data. Given the blurred boundary between reasoning-critical and reasoning-trivial layers, it is plausible that middle layers also contribute to memorization, while late layers are partially involved in reasoning. As a result, excessive weight updates across all layers can lead to overfitting, especially when the training data is synthesized by the model itself. Model merging mitigates this overfitting by limiting weight changes.
\textit{(ii)} The base model retains strong generalization capabilities, while the self-improved model exhibits self-improved reasoning performance. Model merging combines the strengths of both, integrating the generalization ability of the base model with the reasoning improvements from the self-improved model.
\paragraph{Importance-based Iterative Model Merging (IIMM)} We also propose IIMM, which is motivated to aggressively merge the model according to the layer importance as follows:
\begin{equation}
    \boldsymbol{\theta_{m,n}^{t+1}} = \alpha \boldsymbol{\theta_n} + (1-\alpha)(\boldsymbol{\theta_n^{t}} + \frac{NI_n}{{\sum_{n=1}^{N}} I_n}\boldsymbol{\hat{\delta_n^t}}),
\end{equation}
where $n$ denotes $n$-th layer of the model with $N$ layers. However, we find that IIMM is outperformed by IMM because of instability and overfitting datasets for importance score calculation. The detailed experiment and analysis are provided in Appendix~\ref{app:iimm}.

\section{Experiments}

In this section, we conduct extensive experiments to evaluate the effectiveness of our proposed method. Specifically, our experiments aim to address the following research questions: \textit{(i)} Can our method prevent model collapse on complex reasoning tasks during iterative self-improvement? \textit{(ii)} How well does our method perform on OOD reasoning tasks? \textit{(iii)} Can our method be extended from self-improvement to knowledge distillation from a stronger model?

\begin{figure*}[h!]
    \vspace{1pt}
    \centering
    \begin{minipage}[t]{0.33\textwidth}
        \centering
        \includegraphics[width=\linewidth]{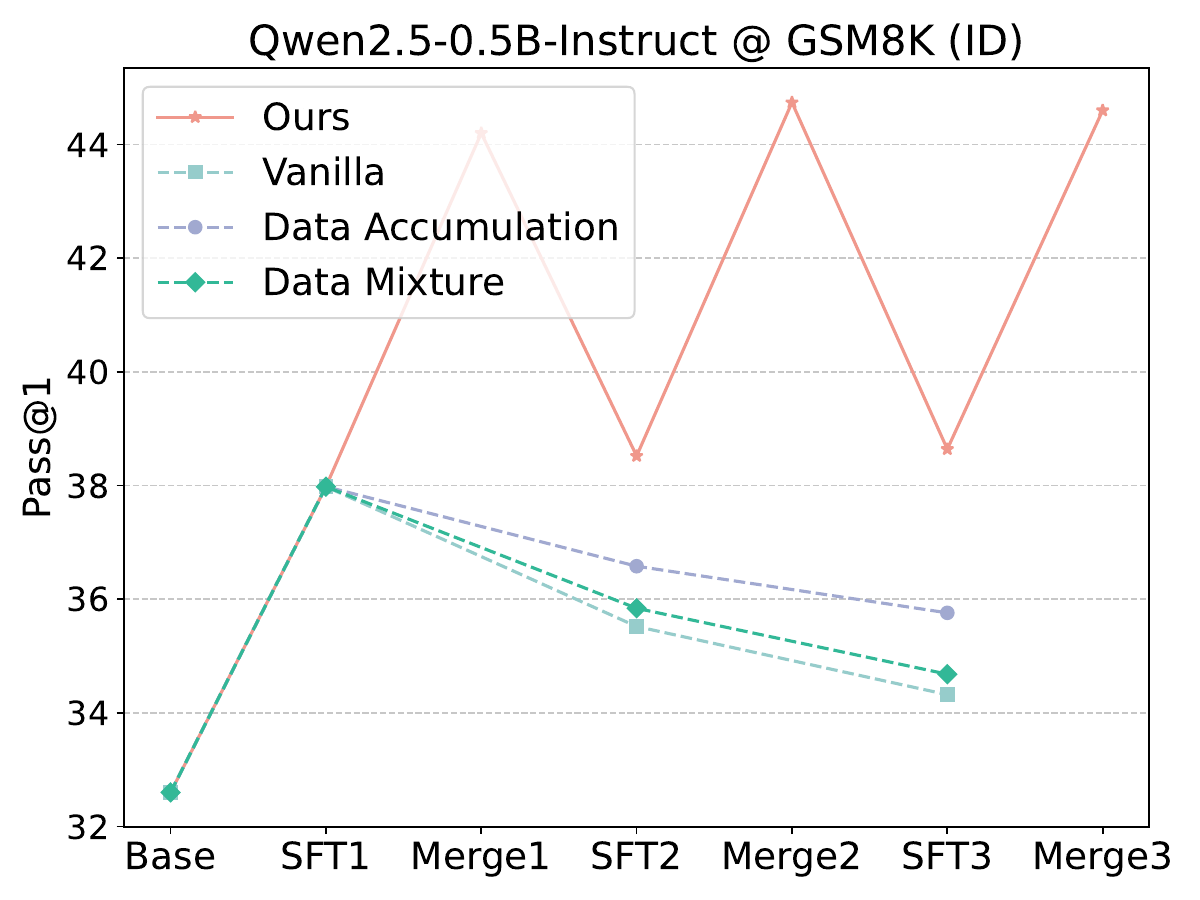}
    \end{minipage}%
    \begin{minipage}[t]{0.33\textwidth}
        \centering
        \includegraphics[width=\linewidth]{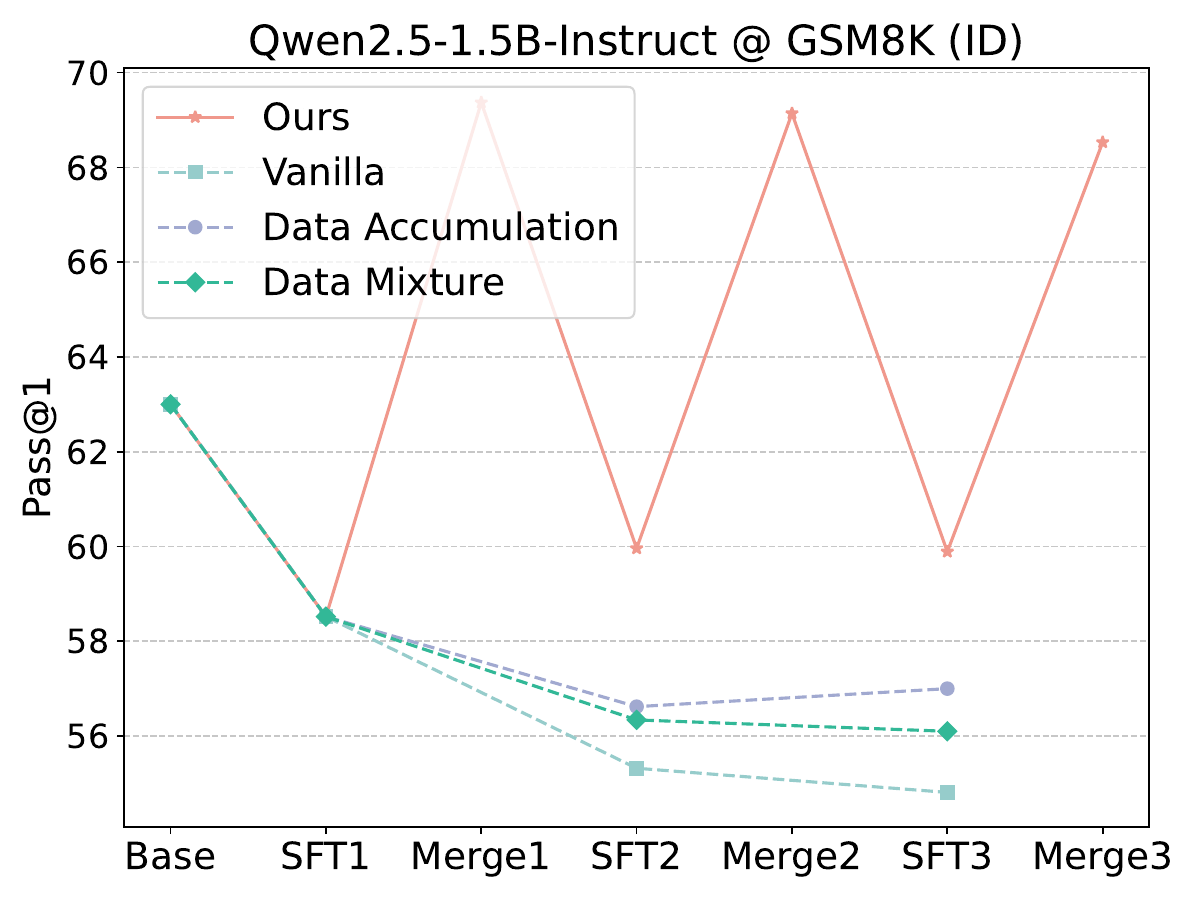}
    \end{minipage}%
    \begin{minipage}[t]{0.33\textwidth}
        \centering
        \includegraphics[width=\linewidth]{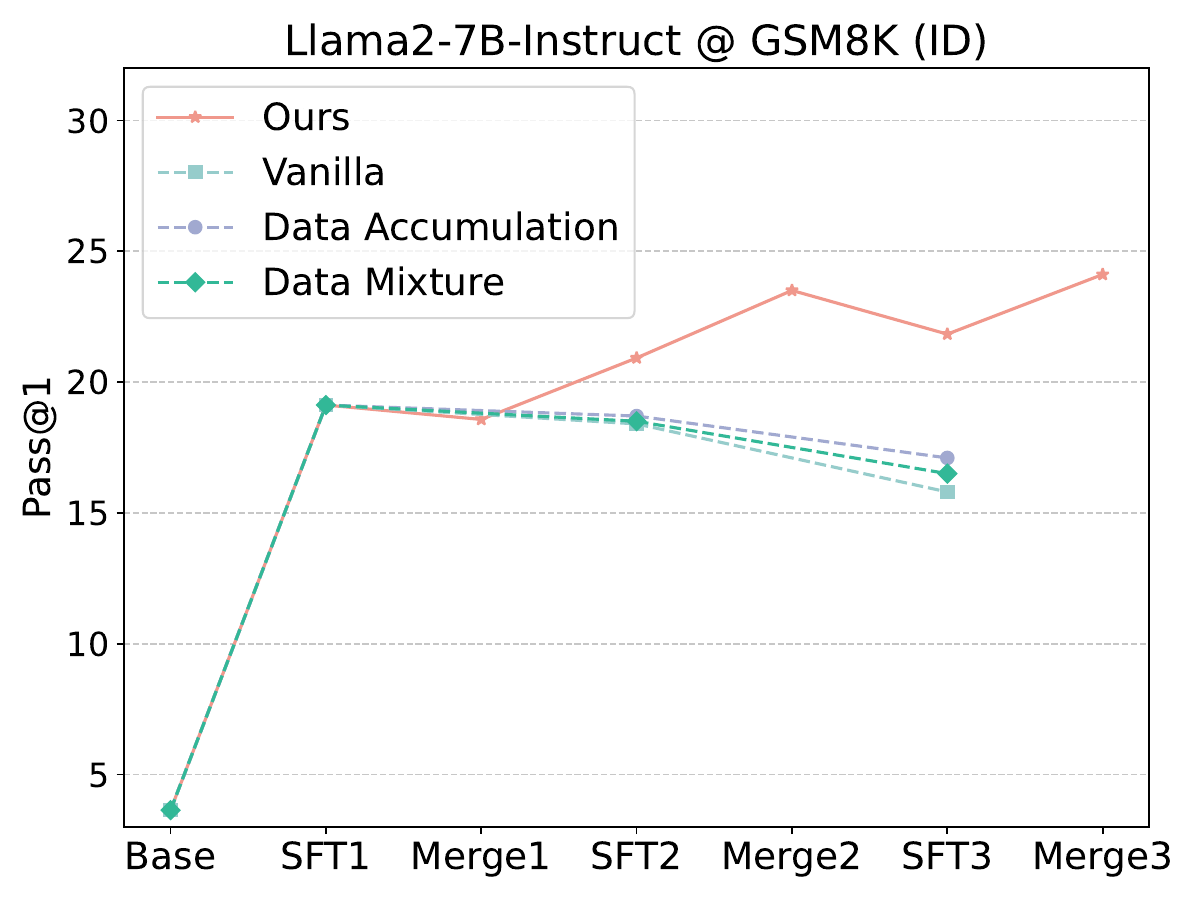}
    \end{minipage}%

\end{figure*}

\begin{figure*}[h!]
    \centering
    \begin{minipage}[t]{0.33\textwidth}
        \centering
        \includegraphics[width=\linewidth]{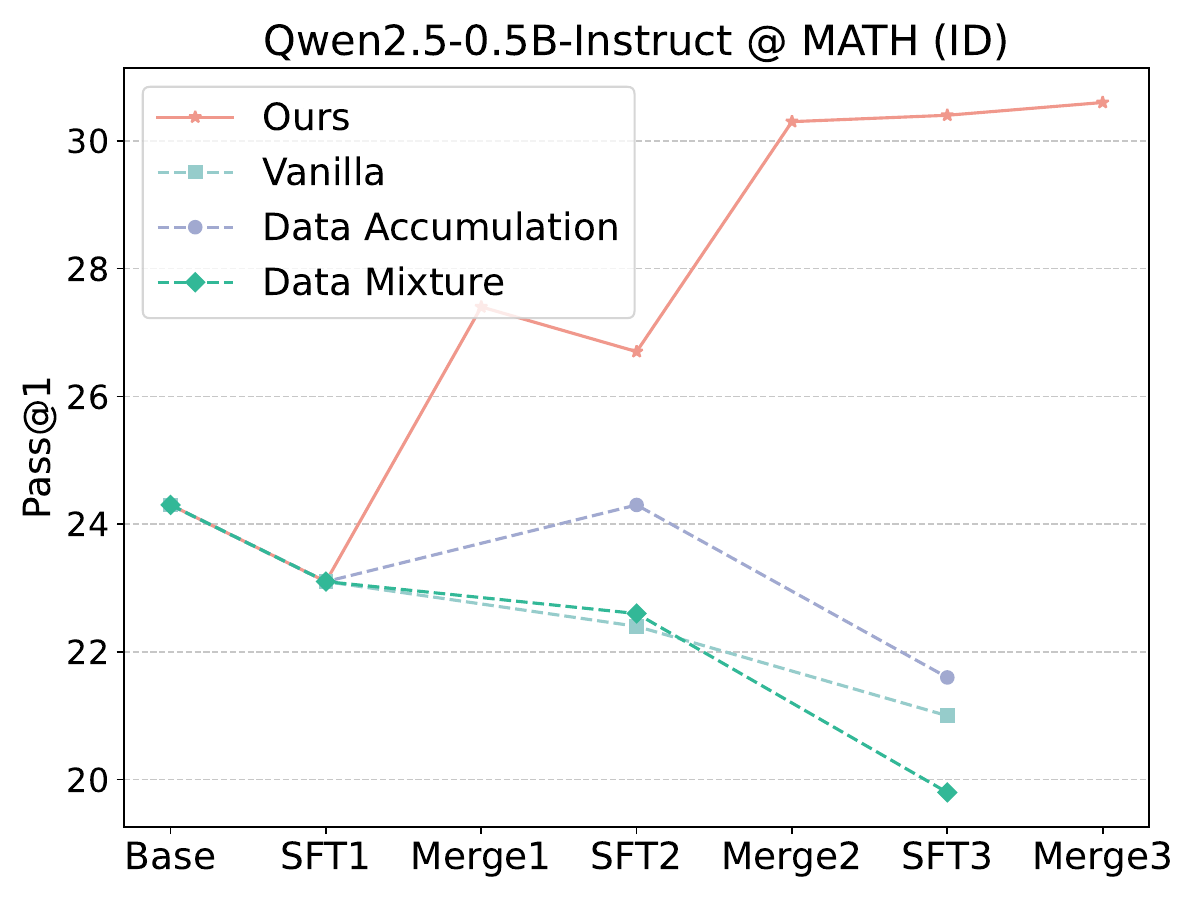}
    \end{minipage}%
    \begin{minipage}[t]{0.33\textwidth}
        \centering
        \includegraphics[width=\linewidth]{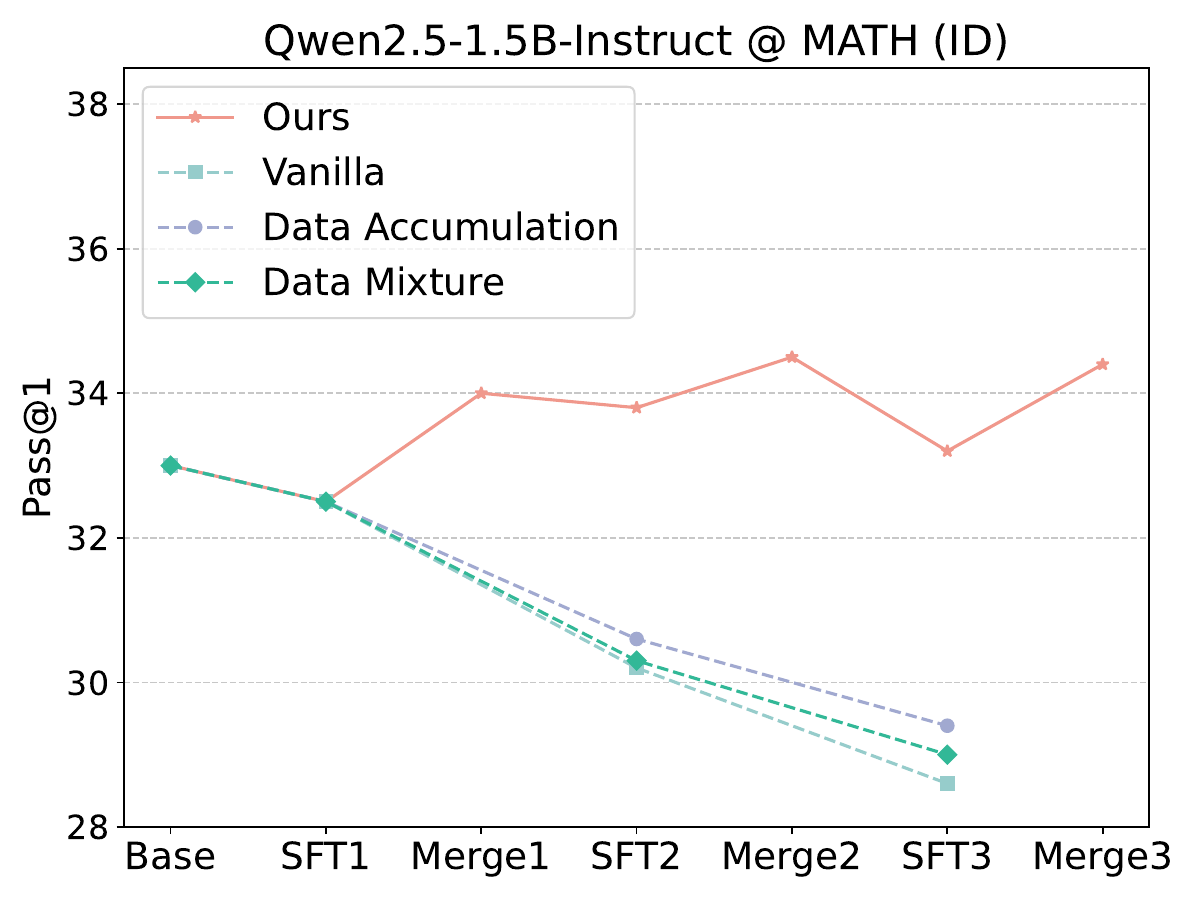}
    \end{minipage}%
    \begin{minipage}[t]{0.33\textwidth}
        \centering
        \includegraphics[width=\linewidth]{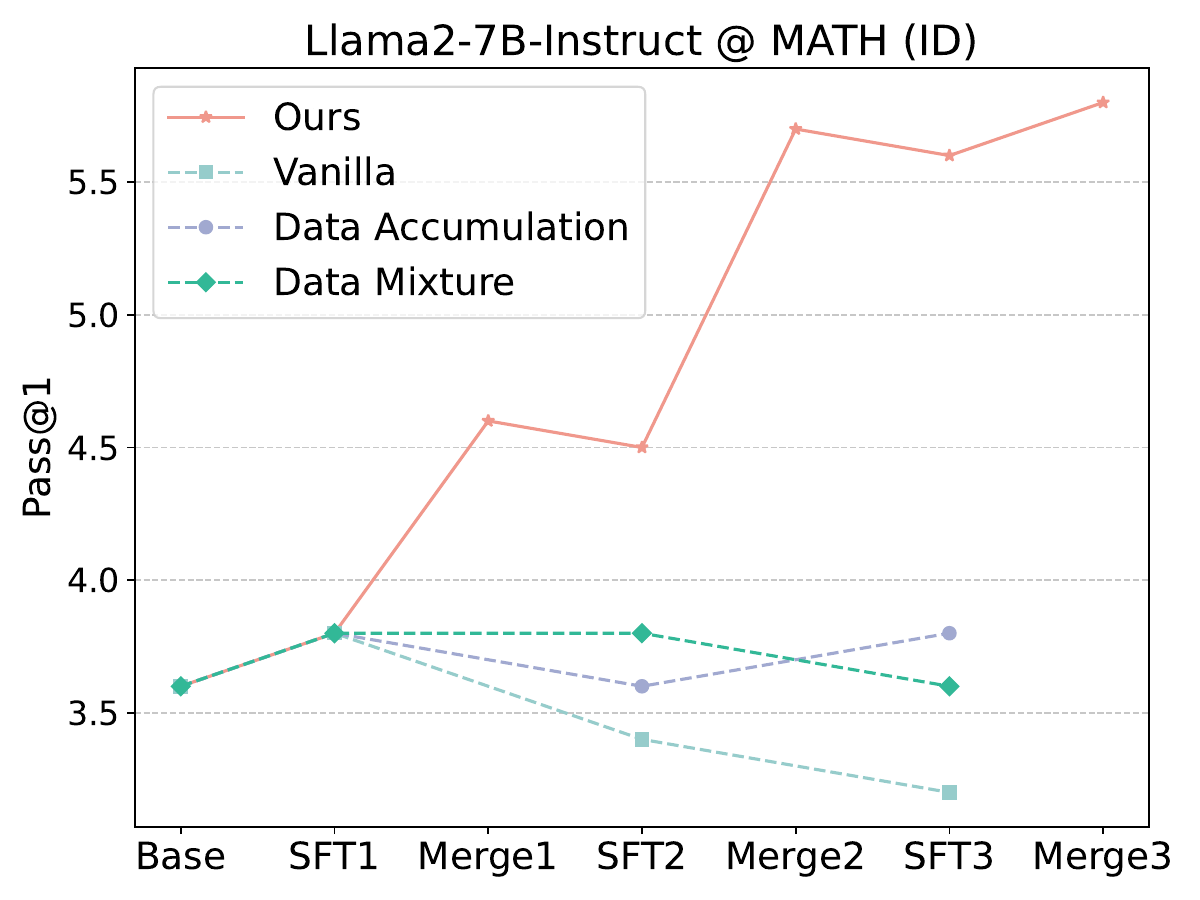}
    \end{minipage}%
    \caption{The model performances on in-domain (ID) datasets. SFT n and Merge n denote the SFT model and merged model in the n-th iteration cycle. The model collapse happens from the first or second iteration for baselines, while our method avoids it and achieves the best performance after model merging.}
    \label{fig:mian_id}

    \label{fig:mian_gsm8k}
\end{figure*}

\begin{figure*}[h!]
    \vspace{1pt}
    \centering
    \begin{minipage}[t]{0.33\textwidth}
        \centering
        \includegraphics[width=\linewidth]{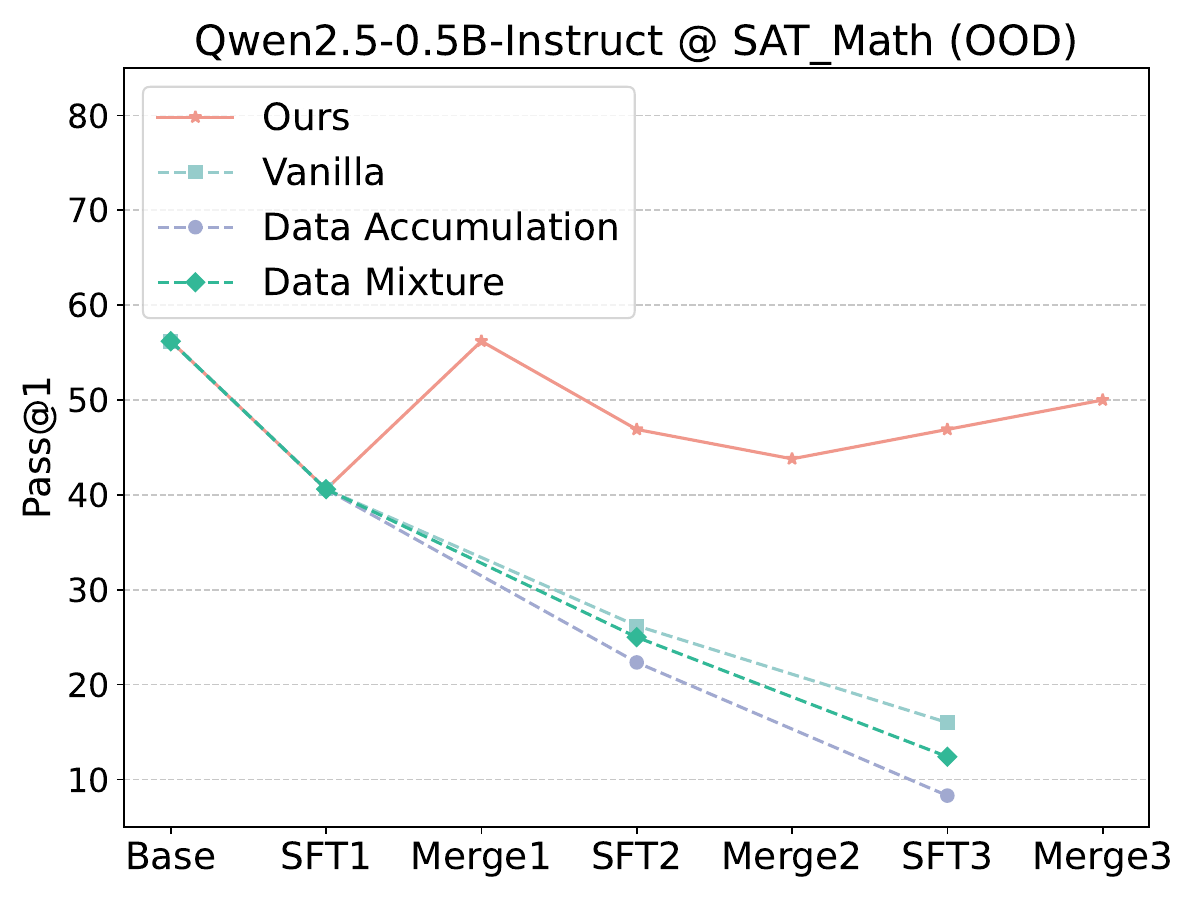}
    \end{minipage}%
    \begin{minipage}[t]{0.33\textwidth}
        \centering
        \includegraphics[width=\linewidth]{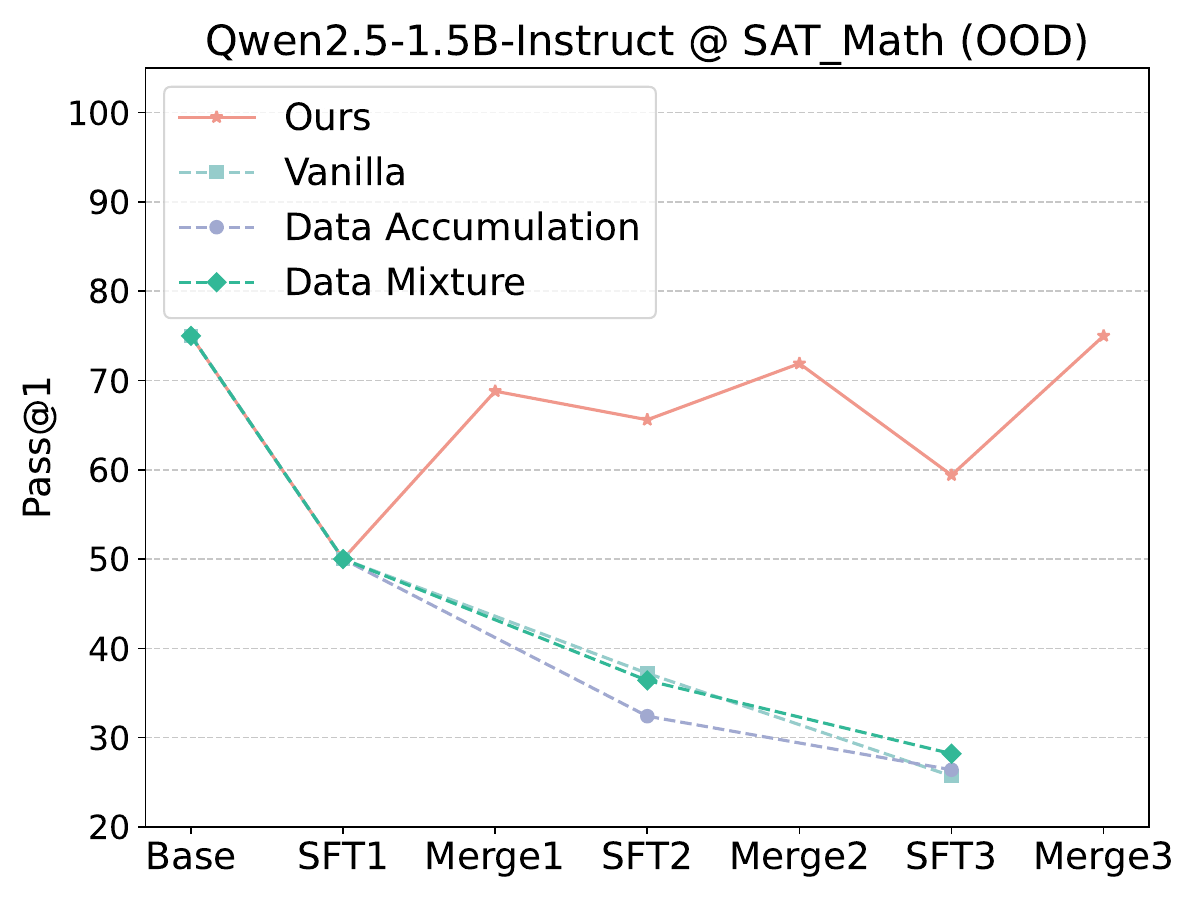}
    \end{minipage}%
    \begin{minipage}[t]{0.33\textwidth}
        \centering
        \includegraphics[width=\linewidth]{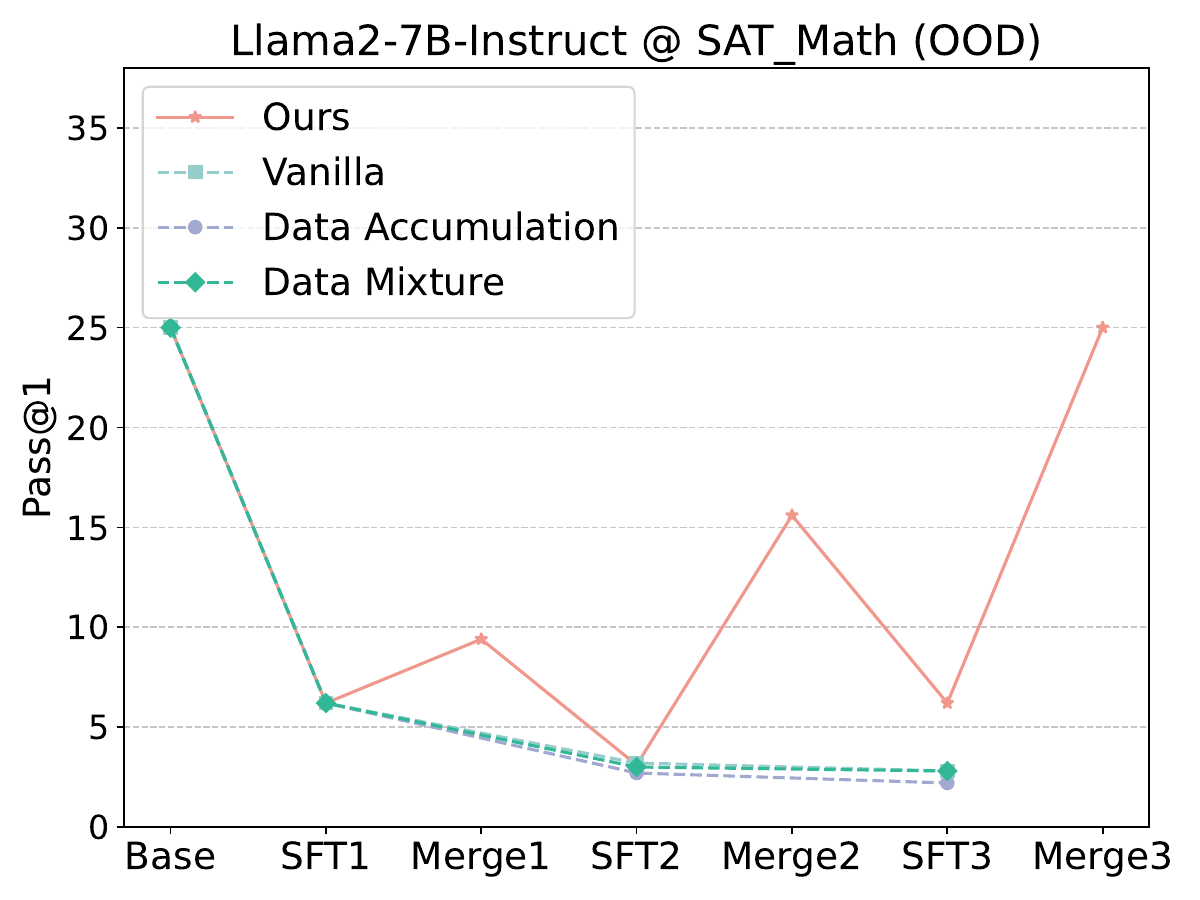}
    \end{minipage}%

    \label{fig:mian_gsm8k_1}
\end{figure*}

\begin{figure*}[h!]
    \vspace{1pt}
    \centering
    \begin{minipage}[t]{0.33\textwidth}
        \centering
        \includegraphics[width=\linewidth]{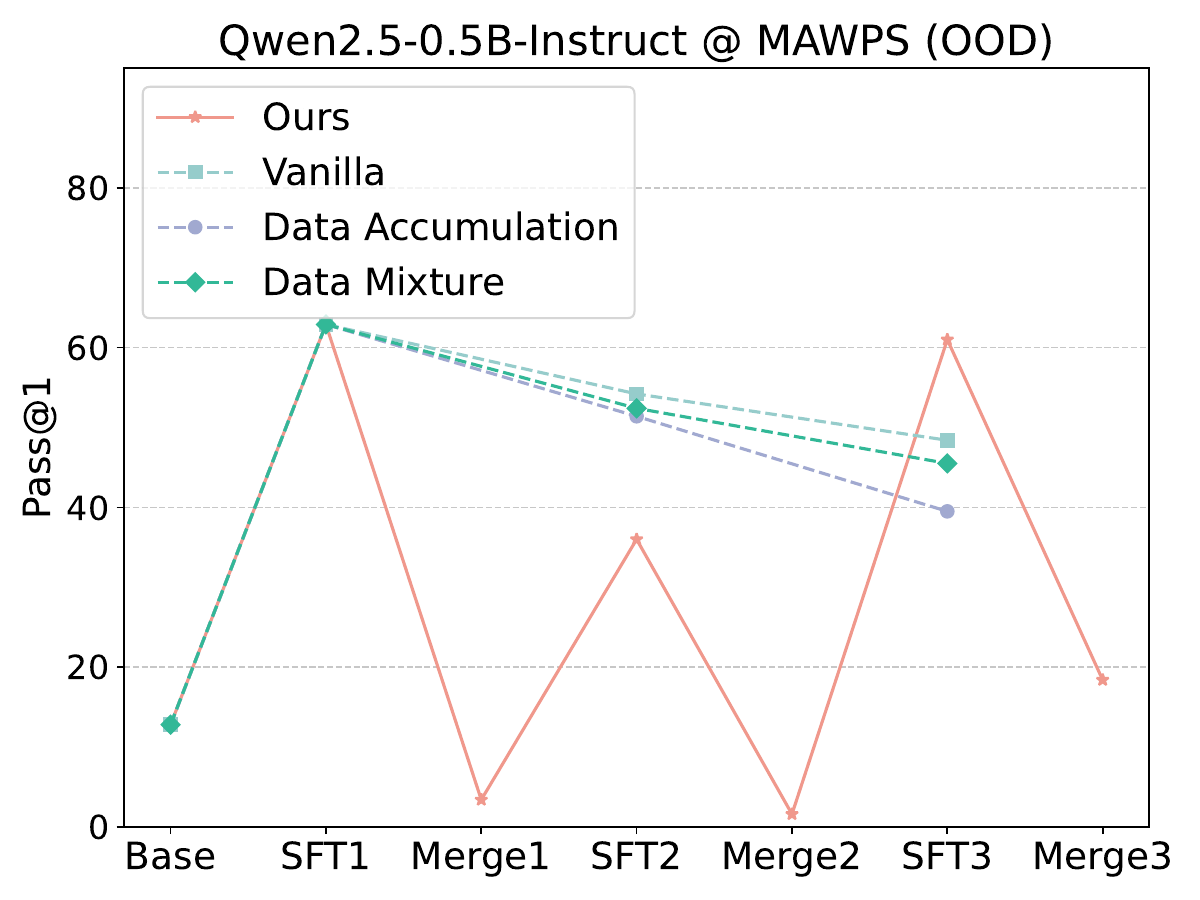}
    \end{minipage}%
    \begin{minipage}[t]{0.33\textwidth}
        \centering
        \includegraphics[width=\linewidth]{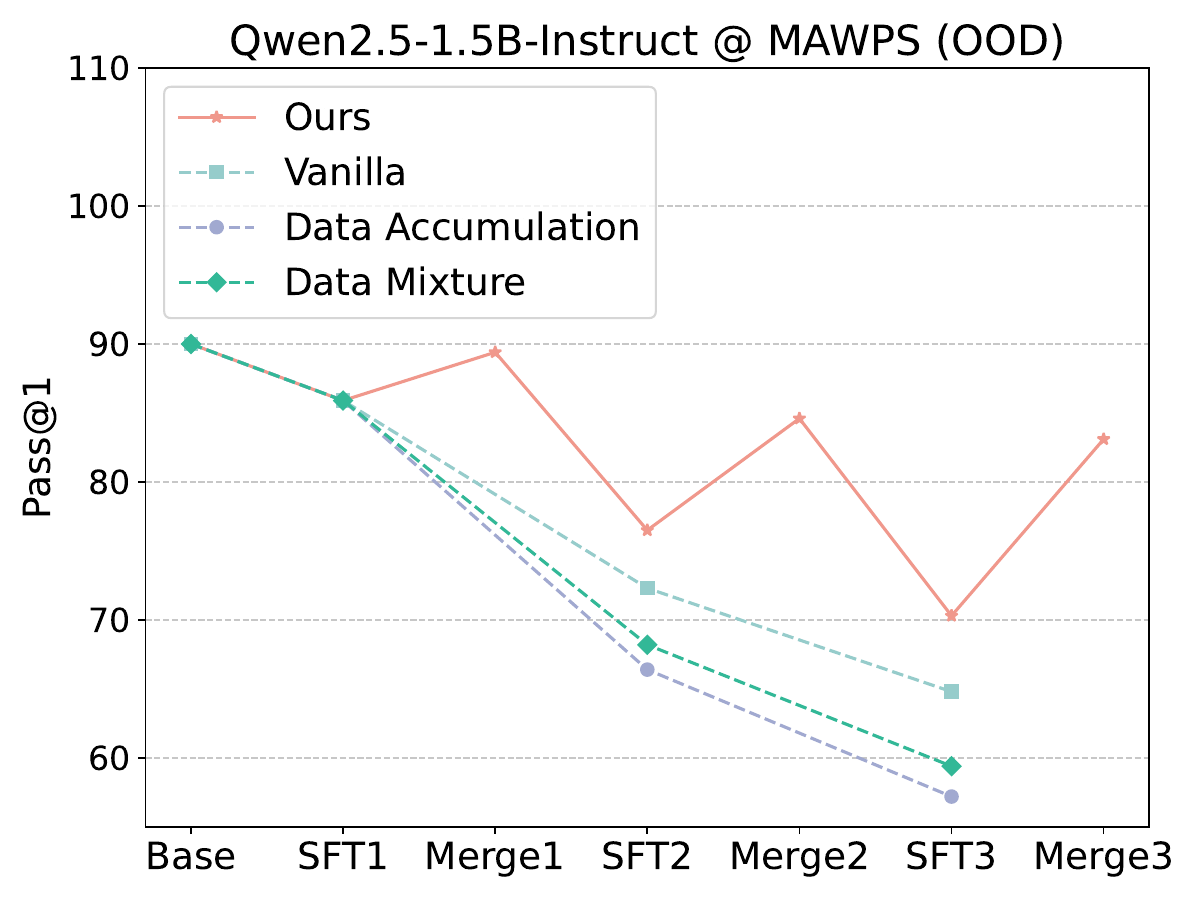}
    \end{minipage}%
    \begin{minipage}[t]{0.33\textwidth}
        \centering
        \includegraphics[width=\linewidth]{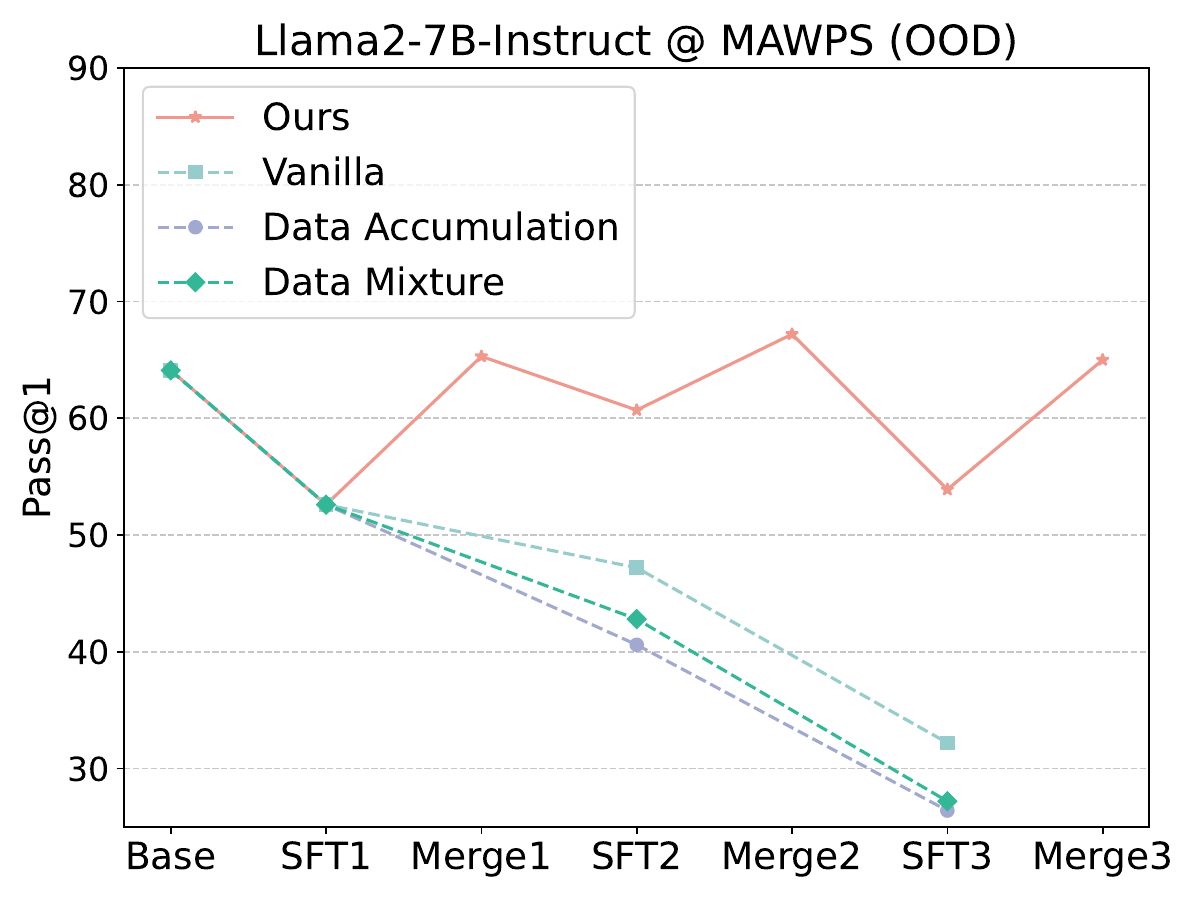}
    \end{minipage}%
    
    \caption{ The model performances on out-of-domain (OOD) datasets. SFT n and Merge n denote the SFT model and merged model in the n-th iteration cycle. Baselines' performances decrease on most datasets, while IMM can generally maintain the OOD performance compared with the original base model.}
    \label{fig:mian_ood}
    \vspace{-5pt}
\end{figure*}

\begin{figure}[h!]

    \centering
    \begin{minipage}[t]{0.24\textwidth}
        \centering
        \includegraphics[width=\linewidth]{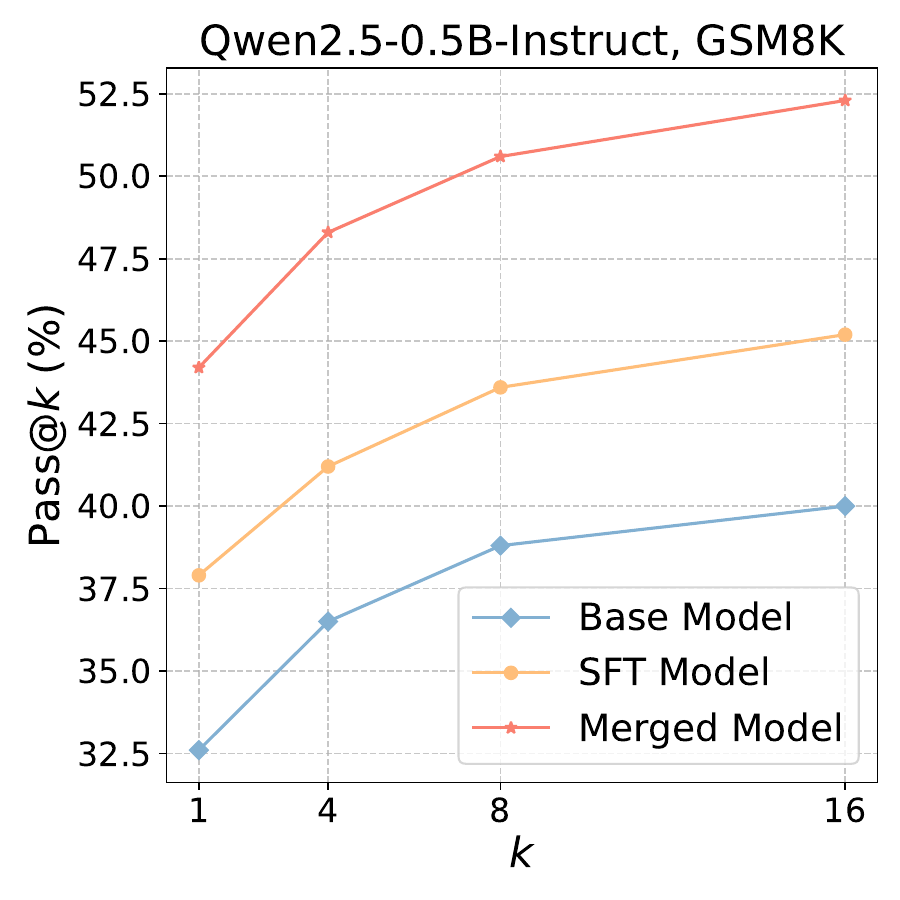}
    \end{minipage}%
    \begin{minipage}[t]{0.24\textwidth}
        \centering
        \includegraphics[width=\linewidth]{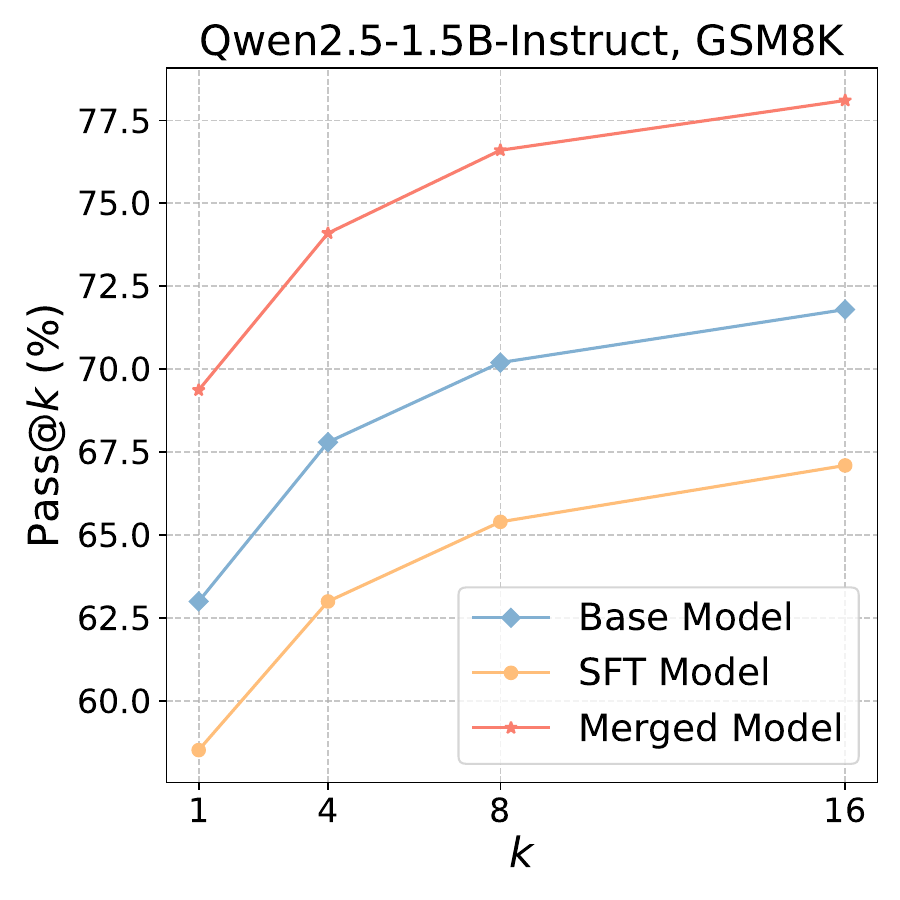}
    \end{minipage}%

    \caption{ID performance with different k for scaling up test-time-computing Pass@k on GSM8K.}
    \label{fig:pass_gsm8k}
     \vspace{-15pt}
\end{figure}

\label{sec:exp}

\subsection{Setup}
\paragraph{Datasets}
We train the model on MATH \cite{hendrycks2021measuring} and GSM-8K \cite{cobbe2021training} datasets correspondingly to evaluate the in-domain reasoning ability of the model, while evaluate it on MAWPS \cite{koncel-kedziorski-etal-2016-mawps}, SAT-Math \cite{zhong-etal-2024-agieval} datasets to evaluate the out-of-domain reasoning ability. 

\paragraph{Models} We include three LLMs at different scales (Qwen2.5-0.5B-Instruct, Qwen2.5-1.5B-Instruct \cite{qwen2.5} and Llama2-7B \cite{touvron2023llama}) for self-improvement training. For the distillation experiments, we include stronger teacher models Qwen2.5-7B-Instruct for distillation. We also provide the recent model Llama3-8B performance in Appendix~\ref{app:llama3}


\paragraph{Baselines}  
We evaluate our method by comparing it with four baselines. First, we consider \textbf{Vanilla} (STaR \cite{zelikman2022star}) , which iteratively generates reasoning data following the procedure in Section \ref{sec:motivation} for self-improvement. Second, we include \textbf{Data Mixture} \cite{shumailov2023curse}, which mitigates performance degradation by mixing a portion of data from previous iterations. Third, we compare with \textbf{Data Accumulation} \cite{gerstgrasser2024is}, which demonstrates that accumulating synthetic data across iterations can prevent model collapse. We also provide a comparison of SFT interventions in Appendix~\ref{app:sft_inter}.

\paragraph{Evaluation}
We evaluate the model performance by computing $\text{pass@}k = \mathbb{E}_{\mathcal{D}_G} \left[ 1 - \frac{\binom{M - c}{k}}{\binom{M}{k}} \right]$,
where \( c \) is the number of correct answers, out of total answer \( M \) and \( \mathbb{E}_{\mathcal{D}_G}[\cdot] \) is the expectation for overall generated dataset \( \mathcal{D}_G \). Therefore, \( \text{pass@}k \) measures the fraction of unique questions that have at least one correct answer when sampling \( k \) answers per question from the model. 

\noindent Additional training and implementation details are provided in Appendix~\ref{app:training_detail}.

\subsection{ID Results with Self-improvement}
\label{exp:main_id}
To answer research question \textit{(i)},
we conducted extensive experiments in a model collapse setting (iterative self-improvement) using two mathematical reasoning datasets, GSM8K and MATH. The results, shown in Figure \ref{fig:mian_id}, highlight that across three self-improvement iterations with three different LLMs, model collapse occurs in the first or second iteration for the baseline methods. In contrast, our method successfully avoids model collapse and achieves the best performance after applying model merging. Not only does our method significantly delay model collapse, but it also maintains superior performance across all iterations. Moreover, we observe that LLMs of all scales benefit from our model merging strategy, with smaller models suffering more severely from model collapse in the absence of this approach.
Given the rising importance of test-time computing \cite{snell2024scaling}, we further evaluate our method by generating multiple answers and measuring pass@$k$ accuracy. As shown in Figure \ref{fig:pass_gsm8k} (more results are presented in Appendix~\ref{app:test_time}), our method consistently improves performance as k increases and outperforms both the base models and the SFT models.

\subsection{OOD Generalization Results}
To answer research question \textit{(ii)},
We evaluate the checkpoints from Section \ref{exp:main_id} using OOD math reasoning datasets: SAT Math and MAWPS. Additional OOD datasets results can be found in Appendix~\ref{app:ood_exp}. The results, presented in Figure \ref{fig:mian_ood}, show that while all other baselines suffer significant OOD performance degradation after iterative self-improvement, our method consistently restores performance after each model merging step and, in some cases, even surpasses the original base model. The only exception is the Qwen2.5-0.5B-Instruct model on the MAWPS dataset. We hypothesize that this dataset closely resembles the in-domain data, where extensive ID training significantly improves performance, which causes a degradation during IMM. We further analyze this unexpected behavior in Appendix~\ref{app:unexpected_behavior}. Overall, these results demonstrate the great potential of our method, as it successfully mitigates the generalization drop commonly observed during SFT.
\subsection{Distillation from Stronger Models}
\label{sec:distill}
Considering self-improvement may be only one of paradigms for LLM distillation, we extend our method to a broader field to answer research question \textit{(iii)}. 
We distill a stronger Qwen2.5-7B-Instruct model into the weaker Qwen2.5-1.5B-Instruct and Llama-2-7B models. The results in Table~\ref{tab:distill} demonstrate that IMM consistently improves or maintains comparable performance on ID tasks, while often achieving significant improvements in OOD performance. This indicates that IMM only preserves task-specific performance but also enhances the model’s generalized reasoning ability when distilling from the teacher model. 


\begin{table}[]

\resizebox{0.48\textwidth}{!}{%
\begin{tabular}{lccccc}
\toprule
\textbf{\begin{tabular}[c]{@{}l@{}}Student\end{tabular}}                  & \textbf{Domain}      & \textbf{Datasets} & \textbf{Base} & \textbf{\begin{tabular}[c]{@{}c@{}}SFT\end{tabular}} & \textbf{Merged} \\ \midrule
\multirow{4}{*}{\begin{tabular}[c]{@{}l@{}}Qwen2.5-\\ 1.5B Instruct\end{tabular}} & \multirow{2}{*}{ID}  & GSM8K             & {\ul 63.0}    & 54.4                                                              & \textbf{71.6}  \\
                                                                                  &                      & MATH              & 24.3          & \textbf{45.0}                                                     & {\ul 42.6}      \\ \cmidrule{3-6} 
                                                                                  & \multirow{2}{*}{OOD} & SAT\_Math         & {\ul 75.0}    & {\ul 75.0}                                                        & \textbf{87.5}   \\
                                                                                  &                      & MAWPS             & \textbf{90.0}   & {\ul 72.8}                                                        & 24.5            \\ \hline
\multirow{4}{*}{Llama2-7B}                                                        & \multirow{2}{*}{ID}  & GSM8K             & 3.6          & \textbf{49.2}                                                     & {\ul 38.8}     \\
                                                                                  &                      & MATH              & 3.6          & {\ul 10.3}                                                        & \textbf{12.5}   \\ \cmidrule{3-6} 
                                                                                  & \multirow{2}{*}{OOD} & SAT\_Math         & {\ul 25.0}      & 18.8                                                              & \textbf{28.1}   \\
                                                                                  &                      & MAWPS             & {\ul 64.1}    & 55.1                                                              & \textbf{76.6}   \\ \bottomrule
\end{tabular}}
\caption{Student models' performance with distilling from stronger model setting. The best and runner-up accuracies are \textbf{bolded} and {\ul underlined} respectively.}
\vspace{-5pt}
\label{tab:distill}
\end{table}

\section{Conclusion}

This study identifies that self-improved LLM reasoners still have the model collapse risk and lack generalized reasoning capability on OOD datasets. Our analysis reveals that the weight changes of layers doesn't match the layer importance. This mismatch suggests that instead of solely learning to reason, the model also memorizes the training data.
To address this issue, we propose the Iterative Model Merge and extensive experiments demonstrate the effectiveness of our method: it not only mitigates model collapse but also make model have generalized reasoning capability.

\section*{Limitations}

The proposed Iterative Model Merging (IMM) method currently employs a fixed-weight merging mechanism between the original and self-improved models. However, more advanced strategies, such as dynamic or layer-adaptive merging, could provide further improvements. Additionally, although IMM has proven to be effective in maintaining generalized reasoning capabilities, it doesn't investigate the strategy of mixing real and synthetic data appropriately, which could further enhance the trade-offs between reasoning improvement and generalization. We leave the exploration of advanced merging mechanisms and the optimal mixture ratio of real and synthetic data for future work. 

\section{Acknowledgement}
This material is based upon work supported by the National Science Foundation under grant No. 2229876 and is supported in part by funds provided by the National Science Foundation, by the Department of Homeland Security, and by IBM. Any opinions, findings, and conclusions or recommendations expressed in this material are those of the author(s) and do not necessarily reflect the views of the National Science Foundation or its federal agency and industry partners.

\bibliography{ref.bib}

\begin{thebibliography}{78}
\providecommand{\natexlab}[1]{#1}

\bibitem[{Alemohammad et~al.(2024)Alemohammad, Casco-Rodriguez, Luzi, Humayun, Babaei, LeJeune, Siahkoohi, and Baraniuk}]{alemohammad2024selfconsuming}
Sina Alemohammad, Josue Casco-Rodriguez, Lorenzo Luzi, Ahmed~Imtiaz Humayun, Hossein Babaei, Daniel LeJeune, Ali Siahkoohi, and Richard Baraniuk. 2024.
\newblock Self-consuming generative models go {MAD}.
\newblock In \emph{International Conference on Learning Representations}.

\bibitem[{Amini et~al.(2019)Amini, Gabriel, Lin, Koncel-Kedziorski, Choi, and Hajishirzi}]{amini2019mathqa}
Aida Amini, Saadia Gabriel, Peter Lin, Rik Koncel-Kedziorski, Yejin Choi, and Hannaneh Hajishirzi. 2019.
\newblock Mathqa: Towards interpretable math word problem solving with operation-based formalisms.
\newblock \emph{arXiv preprint arXiv:1905.13319}.

\bibitem[{Bansal et~al.(2024)Bansal, Hosseini, Agarwal, Tran, and Kazemi}]{bansal2024smaller}
Hritik Bansal, Arian Hosseini, Rishabh Agarwal, Vinh~Q Tran, and Mehran Kazemi. 2024.
\newblock Smaller, weaker, yet better: Training llm reasoners via compute-optimal sampling.
\newblock \emph{arXiv preprint arXiv:2408.16737}.

\bibitem[{Bertrand et~al.(2023)Bertrand, Bose, Duplessis, Jiralerspong, and Gidel}]{bertrand2023stability}
Quentin Bertrand, Avishek~Joey Bose, Alexandre Duplessis, Marco Jiralerspong, and Gauthier Gidel. 2023.
\newblock On the stability of iterative retraining of generative models on their own data.
\newblock \emph{arXiv preprint arXiv:2310.00429}.

\bibitem[{Bertrand et~al.(2024)Bertrand, Bose, Duplessis, Jiralerspong, and Gidel}]{bertrand2024on}
Quentin Bertrand, Joey Bose, Alexandre Duplessis, Marco Jiralerspong, and Gauthier Gidel. 2024.
\newblock On the stability of iterative retraining of generative models on their own data.
\newblock In \emph{The Twelfth International Conference on Learning Representations}.

\bibitem[{Bohacek and Farid(2023)}]{bohacek2023nepotistically}
Matyas Bohacek and Hany Farid. 2023.
\newblock Nepotistically trained generative-ai models collapse.
\newblock \emph{arXiv preprint arXiv:2311.12202}.

\bibitem[{Cobbe et~al.(2021)Cobbe, Kosaraju, Bavarian, Chen, Jun, Kaiser, Plappert, Tworek, Hilton, Nakano et~al.}]{cobbe2021training}
Karl Cobbe, Vineet Kosaraju, Mohammad Bavarian, Mark Chen, Heewoo Jun, Lukasz Kaiser, Matthias Plappert, Jerry Tworek, Jacob Hilton, Reiichiro Nakano, et~al. 2021.
\newblock Training verifiers to solve math word problems, 2021.
\newblock \emph{URL https://arxiv. org/abs/2110.14168}.

\bibitem[{{DeepMind}(2024{\natexlab{a}})}]{deepmind2023imo}
{DeepMind}. 2024{\natexlab{a}}.
\newblock Ai solves imo problems at a silver medal level.
\newblock \emph{URL https://deepmind.google/discover/blog/ai-solves-imo-problems-at-silver-medal-level/}.

\bibitem[{{DeepMind}(2024{\natexlab{b}})}]{deepmind2023alphazero}
{DeepMind}. 2024{\natexlab{b}}.
\newblock Alphazero: Shedding new light on chess, shogi, and go.
\newblock \emph{URL https://deepmind.google/discover/blog/alphazero-shedding-new-light-on-chess-shogi-and-go/}.

\bibitem[{Diao et~al.(2025)Diao, Zhang, Kong, Wu, Ma, Ouyang, Qing, Vosoughi, and Gui}]{diao2025soundmind}
Xingjian Diao, Chunhui Zhang, Keyi Kong, Weiyi Wu, Chiyu Ma, Zhongyu Ouyang, Peijun Qing, Soroush Vosoughi, and Jiang Gui. 2025.
\newblock Soundmind: Rl-incentivized logic reasoning for audio-language models.
\newblock \emph{arXiv preprint arXiv:2506.12935}.

\bibitem[{Dohmatob et~al.(2025)Dohmatob, Feng, Subramonian, and Kempe}]{dohmatob2025strong}
Elvis Dohmatob, Yunzhen Feng, Arjun Subramonian, and Julia Kempe. 2025.
\newblock Strong model collapse.
\newblock In \emph{The Thirteenth International Conference on Learning Representations}.

\bibitem[{Feng et~al.(2025)Feng, Dohmatob, Yang, Charton, and Kempe}]{feng2025beyond}
Yunzhen Feng, Elvis Dohmatob, Pu~Yang, Francois Charton, and Julia Kempe. 2025.
\newblock Beyond model collapse: Scaling up with synthesized data requires verification.
\newblock In \emph{The Thirteenth International Conference on Learning Representations}.

\bibitem[{Fu et~al.(2024{\natexlab{a}})Fu, Zhang, Wang, Tian, and Tao}]{fu2024towards}
Shi Fu, Sen Zhang, Yingjie Wang, Xinmei Tian, and Dacheng Tao. 2024{\natexlab{a}}.
\newblock Towards theoretical understandings of self-consuming generative models.
\newblock In \emph{Forty-first International Conference on Machine Learning}.

\bibitem[{Fu et~al.(2024{\natexlab{b}})Fu, Yu, Li, Qian, Zhang, Yuan, Shi, Yakunin, and Lin}]{fu2024amoeballm}
Yonggan Fu, Zhongzhi Yu, Junwei Li, Jiayi Qian, Yongan Zhang, Xiangchi Yuan, Dachuan Shi, Roman Yakunin, and Yingyan~Celine Lin. 2024{\natexlab{b}}.
\newblock Amoeballm: Constructing any-shape large language models for efficient and instant deployment.
\newblock \emph{Advances in Neural Information Processing Systems}, 37:78299--78319.

\bibitem[{Gerstgrasser et~al.(2024)Gerstgrasser, Schaeffer, Dey, Rafailov, Korbak, Sleight, Agrawal, Hughes, Pai, Gromov, Roberts, Yang, Donoho, and Koyejo}]{gerstgrasser2024is}
Matthias Gerstgrasser, Rylan Schaeffer, Apratim Dey, Rafael Rafailov, Tomasz Korbak, Henry Sleight, Rajashree Agrawal, John Hughes, Dhruv~Bhandarkar Pai, Andrey Gromov, Dan Roberts, Diyi Yang, David~L. Donoho, and Sanmi Koyejo. 2024.
\newblock Is model collapse inevitable? breaking the curse of recursion by accumulating real and synthetic data.
\newblock In \emph{First Conference on Language Modeling}.

\bibitem[{Gillman et~al.(2024)Gillman, Freeman, Aggarwal, Hsu, Luo, Tian, and Sun}]{gillman2024selfcorrecting}
Nate Gillman, Michael Freeman, Daksh Aggarwal, Chia-Hong Hsu, Calvin Luo, Yonglong Tian, and Chen Sun. 2024.
\newblock Self-correcting self-consuming loops for generative model training.
\newblock In \emph{International Conference on Machine Learning}.

\bibitem[{Gulcehre et~al.(2023)Gulcehre, Paine, Srinivasan, Konyushkova, Weerts, Sharma, Siddhant, Ahern, Wang, Gu et~al.}]{gulcehre2023reinforced}
Caglar Gulcehre, Tom~Le Paine, Srivatsan Srinivasan, Ksenia Konyushkova, Lotte Weerts, Abhishek Sharma, Aditya Siddhant, Alex Ahern, Miaosen Wang, Chenjie Gu, et~al. 2023.
\newblock Reinforced self-training (rest) for language modeling.
\newblock \emph{arXiv preprint arXiv:2308.08998}.

\bibitem[{Guo et~al.(2025)Guo, Yang, Zhang, Song, Zhang, Xu, Zhu, Ma, Wang, Bi et~al.}]{guo2025deepseek}
Daya Guo, Dejian Yang, Haowei Zhang, Junxiao Song, Ruoyu Zhang, Runxin Xu, Qihao Zhu, Shirong Ma, Peiyi Wang, Xiao Bi, et~al. 2025.
\newblock Deepseek-r1: Incentivizing reasoning capability in llms via reinforcement learning.
\newblock \emph{arXiv preprint arXiv:2501.12948}.

\bibitem[{Guo et~al.(2024{\natexlab{a}})Guo, Xu, and Ritter}]{guo-etal-2024-meta}
Ruohao Guo, Wei Xu, and Alan Ritter. 2024{\natexlab{a}}.
\newblock Meta-tuning {LLM}s to leverage lexical knowledge for generalizable language style understanding.
\newblock In \emph{Proceedings of the 62nd Annual Meeting of the Association for Computational Linguistics}.

\bibitem[{Guo et~al.(2023)Guo, Shang, Vazirgiannis, and Clavel}]{guo2023curious}
Yanzhu Guo, Guokan Shang, Michalis Vazirgiannis, and Chlo{\'e} Clavel. 2023.
\newblock The curious decline of linguistic diversity: Training language models on synthetic text.
\newblock \emph{arXiv preprint arXiv:2311.09807}.

\bibitem[{Guo et~al.(2024{\natexlab{b}})Guo, Shang, Vazirgiannis, and Clavel}]{guo-etal-2024-curious}
Yanzhu Guo, Guokan Shang, Michalis Vazirgiannis, and Chlo{\'e} Clavel. 2024{\natexlab{b}}.
\newblock The curious decline of linguistic diversity: Training language models on synthetic text.
\newblock In \emph{Findings of the Association for Computational Linguistics: NAACL 2024}.

\bibitem[{Hataya et~al.(2023)Hataya, Bao, and Arai}]{hataya2023will}
Ryuichiro Hataya, Han Bao, and Hiromi Arai. 2023.
\newblock Will large-scale generative models corrupt future datasets?
\newblock In \emph{Proceedings of the IEEE/CVF International Conference on Computer Vision}.

\bibitem[{He et~al.(2022)He, Sun, Yu, Xue, Zhang, Torr, Bai, and Qi}]{he2022synthetic}
Ruifei He, Shuyang Sun, Xin Yu, Chuhui Xue, Wenqing Zhang, Philip Torr, Song Bai, and Xiaojuan Qi. 2022.
\newblock Is synthetic data from generative models ready for image recognition?
\newblock \emph{arXiv preprint arXiv:2210.07574}.

\bibitem[{Hendrycks et~al.(2020)Hendrycks, Burns, Basart, Zou, Mazeika, Song, and Steinhardt}]{hendrycks2020measuring}
Dan Hendrycks, Collin Burns, Steven Basart, Andy Zou, Mantas Mazeika, Dawn Song, and Jacob Steinhardt. 2020.
\newblock Measuring massive multitask language understanding.
\newblock \emph{arXiv preprint arXiv:2009.03300}.

\bibitem[{Hendrycks et~al.(2021)Hendrycks, Burns, Kadavath, Arora, Basart, Tang, Song, and Steinhardt}]{hendrycks2021measuring}
Dan Hendrycks, Collin Burns, Saurav Kadavath, Akul Arora, Steven Basart, Eric Tang, Dawn Song, and Jacob Steinhardt. 2021.
\newblock Measuring mathematical problem solving with the {MATH} dataset.
\newblock In \emph{Neural Information Processing Systems Datasets and Benchmarks Track}.

\bibitem[{Hosseini et~al.(2024)Hosseini, Yuan, Malkin, Courville, Sordoni, and Agarwal}]{hosseini2024v}
Arian Hosseini, Xingdi Yuan, Nikolay Malkin, Aaron Courville, Alessandro Sordoni, and Rishabh Agarwal. 2024.
\newblock V-star: Training verifiers for self-taught reasoners.
\newblock \emph{arXiv preprint arXiv:2402.06457}.

\bibitem[{Hu et~al.(2024)Hu, Li, Xie, Jiang, Stoica, Jin, and Zhang}]{hu2024gamearena}
Lanxiang Hu, Qiyu Li, Anze Xie, Nan Jiang, Ion Stoica, Haojian Jin, and Hao Zhang. 2024.
\newblock Gamearena: Evaluating llm reasoning through live computer games.
\newblock \emph{arXiv preprint arXiv:2412.06394}.

\bibitem[{Huang et~al.(2022)Huang, Gu, Hou, Wu, Wang, Yu, and Han}]{huang2022large}
Jiaxin Huang, Shixiang~Shane Gu, Le~Hou, Yuexin Wu, Xuezhi Wang, Hongkun Yu, and Jiawei Han. 2022.
\newblock Large language models can self-improve.
\newblock \emph{arXiv preprint arXiv:2210.11610}.

\bibitem[{Jaech et~al.(2024)Jaech, Kalai, Lerer, Richardson, El-Kishky, Low, Helyar, Madry, Beutel, Carney et~al.}]{jaech2024openai}
Aaron Jaech, Adam Kalai, Adam Lerer, Adam Richardson, Ahmed El-Kishky, Aiden Low, Alec Helyar, Aleksander Madry, Alex Beutel, Alex Carney, et~al. 2024.
\newblock Openai o1 system card.
\newblock \emph{arXiv preprint arXiv:2412.16720}.

\bibitem[{Jin et~al.(2024)Jin, Che, Peng, Li, Metaxas, and Pavone}]{jin2024learning}
Can Jin, Tong Che, Hongwu Peng, Yiyuan Li, Dimitris Metaxas, and Marco Pavone. 2024.
\newblock Learning from teaching regularization: Generalizable correlations should be easy to imitate.
\newblock \emph{Advances in Neural Information Processing Systems}, 37:966--994.

\bibitem[{Jin et~al.(2025)Jin, Peng, Zhang, Tang, Metaxas, and Che}]{jin2025two}
Can Jin, Hongwu Peng, Qixin Zhang, Yujin Tang, Dimitris~N Metaxas, and Tong Che. 2025.
\newblock Two heads are better than one: Test-time scaling of multi-agent collaborative reasoning.
\newblock \emph{arXiv preprint arXiv:2504.09772}.

\bibitem[{Kojima et~al.(2022)Kojima, Gu, Reid, Matsuo, and Iwasawa}]{kojima2022large}
Takeshi Kojima, Shixiang~Shane Gu, Machel Reid, Yutaka Matsuo, and Yusuke Iwasawa. 2022.
\newblock Large language models are zero-shot reasoners.
\newblock \emph{Advances in neural information processing systems}.

\bibitem[{Koncel-Kedziorski et~al.(2016)Koncel-Kedziorski, Roy, Amini, Kushman, and Hajishirzi}]{koncel-kedziorski-etal-2016-mawps}
Rik Koncel-Kedziorski, Subhro Roy, Aida Amini, Nate Kushman, and Hannaneh Hajishirzi. 2016.
\newblock {MAWPS}: A math word problem repository.
\newblock In \emph{Proceedings of the 2016 Conference of the North {A}merican Chapter of the Association for Computational Linguistics: Human Language Technologies}.

\bibitem[{Kwon et~al.(2023)Kwon, Li, Zhuang, Sheng, Zheng, Yu, Gonzalez, Zhang, and Stoica}]{kwon2023efficient}
Woosuk Kwon, Zhuohan Li, Siyuan Zhuang, Ying Sheng, Lianmin Zheng, Cody~Hao Yu, Joseph~E. Gonzalez, Hao Zhang, and Ion Stoica. 2023.
\newblock Efficient memory management for large language model serving with pagedattention.
\newblock In \emph{Proceedings of the ACM SIGOPS 29th Symposium on Operating Systems Principles}.

\bibitem[{Li et~al.(2022)Li, Choi, Chung, Kushman, Schrittwieser, Leblond, Eccles, Keeling, Gimeno, Dal~Lago et~al.}]{li2022competition}
Yujia Li, David Choi, Junyoung Chung, Nate Kushman, Julian Schrittwieser, R{\'e}mi Leblond, Tom Eccles, James Keeling, Felix Gimeno, Agustin Dal~Lago, et~al. 2022.
\newblock Competition-level code generation with alphacode.
\newblock \emph{Science}.

\bibitem[{Li et~al.(2024)Li, Jiang, Xie, Song, Lian, and Wei}]{li2024understanding}
Zhaoyi Li, Gangwei Jiang, Hong Xie, Linqi Song, Defu Lian, and Ying Wei. 2024.
\newblock Understanding and patching compositional reasoning in llms.
\newblock \emph{Findings of the Association for Computational Linguistics}.

\bibitem[{Lightman et~al.(2023)Lightman, Kosaraju, Burda, Edwards, Baker, Lee, Leike, Schulman, Sutskever, and Cobbe}]{lightman2023let}
Hunter Lightman, Vineet Kosaraju, Yura Burda, Harri Edwards, Bowen Baker, Teddy Lee, Jan Leike, John Schulman, Ilya Sutskever, and Karl Cobbe. 2023.
\newblock Let's verify step by step.
\newblock \emph{arXiv preprint arXiv:2305.20050}.

\bibitem[{Liu et~al.(2025)Liu, Dou, Yuan, Zhang, Tan, and Jiang}]{liu2025modality}
Zheyuan Liu, Guangyao Dou, Xiangchi Yuan, Chunhui Zhang, Zhaoxuan Tan, and Meng Jiang. 2025.
\newblock Modality-aware neuron pruning for unlearning in multimodal large language models.
\newblock \emph{The 63rd Annual Meeting of the Association for Computational Linguistics}.

\bibitem[{Lu et~al.(2024)Lu, Zhou, Ren, Wang, Shi, Pan, Zhan, and Li}]{lu2024mathgenie}
Zimu Lu, Aojun Zhou, Houxing Ren, Ke~Wang, Weikang Shi, Junting Pan, Mingjie Zhan, and Hongsheng Li. 2024.
\newblock Mathgenie: Generating synthetic data with question back-translation for enhancing mathematical reasoning of llms.
\newblock \emph{arXiv preprint arXiv:2402.16352}.

\bibitem[{Ma et~al.(2023)Ma, Fang, and Wang}]{ma2023llm}
Xinyin Ma, Gongfan Fang, and Xinchao Wang. 2023.
\newblock Llm-pruner: On the structural pruning of large language models.
\newblock \emph{Advances in neural information processing systems}, 36:21702--21720.

\bibitem[{Merchant et~al.(2020)Merchant, Rahimtoroghi, Pavlick, and Tenney}]{merchant-etal-2020-happens}
Amil Merchant, Elahe Rahimtoroghi, Ellie Pavlick, and Ian Tenney. 2020.
\newblock \href {https://aclanthology.org/2020.blackboxnlp-1.4/} {What happens to {BERT} embeddings during fine-tuning?}
\newblock In \emph{Proceedings of the Third BlackboxNLP Workshop on Analyzing and Interpreting Neural Networks for NLP}.

\bibitem[{Miao et~al.(2020)Miao, Liang, and Su}]{miao-etal-2020-diverse}
Shen-yun Miao, Chao-Chun Liang, and Keh-Yih Su. 2020.
\newblock A diverse corpus for evaluating and developing {E}nglish math word problem solvers.
\newblock In \emph{Proceedings of the 58th Annual Meeting of the Association for Computational Linguistics}.

\bibitem[{Mosbach et~al.(2020)Mosbach, Khokhlova, Hedderich, and Klakow}]{mosbach-etal-2020-interplay}
Marius Mosbach, Anna Khokhlova, Michael~A. Hedderich, and Dietrich Klakow. 2020.
\newblock On the interplay between fine-tuning and sentence-level probing for linguistic knowledge in pre-trained transformers.
\newblock In \emph{Proceedings of the Third BlackboxNLP Workshop on Analyzing and Interpreting Neural Networks for NLP}.

\bibitem[{OpenAI(2025)}]{openai2025o3}
OpenAI. 2025.
\newblock Openai o3-mini.
\newblock \emph{URL https://openai.com/index/openai-o3-mini/}.

\bibitem[{Padmakumar and He(2024)}]{padmakumar2024does}
Vishakh Padmakumar and He~He. 2024.
\newblock Does writing with language models reduce content diversity?
\newblock In \emph{The Twelfth International Conference on Learning Representations}.

\bibitem[{Pang et~al.(2024)Pang, Yuan, He, Cho, Sukhbaatar, and Weston}]{pang2024iterative}
Richard~Yuanzhe Pang, Weizhe Yuan, He~He, Kyunghyun Cho, Sainbayar Sukhbaatar, and Jason~E Weston. 2024.
\newblock Iterative reasoning preference optimization.
\newblock In \emph{The Thirty-eighth Annual Conference on Neural Information Processing Systems}.

\bibitem[{Patel et~al.(2021)Patel, Bhattamishra, and Goyal}]{patel-etal-2021-nlp}
Arkil Patel, Satwik Bhattamishra, and Navin Goyal. 2021.
\newblock Are {NLP} models really able to solve simple math word problems?
\newblock In \emph{Proceedings of the 2021 Conference of the North American Chapter of the Association for Computational Linguistics: Human Language Technologies}.

\bibitem[{Prasad et~al.(2024)Prasad, Yuan, Pang, Xu, Fazel-Zarandi, Bansal, Sukhbaatar, Weston, and Yu}]{prasad2024self}
Archiki Prasad, Weizhe Yuan, Richard~Yuanzhe Pang, Jing Xu, Maryam Fazel-Zarandi, Mohit Bansal, Sainbayar Sukhbaatar, Jason Weston, and Jane Yu. 2024.
\newblock Self-consistency preference optimization.
\newblock \emph{arXiv preprint arXiv:2411.04109}.

\bibitem[{Qi et~al.(2025)Qi, MA, Xu, Zhang, Yang, and Yang}]{qi2025mutual}
Zhenting Qi, Mingyuan MA, Jiahang Xu, Li~Lyna Zhang, Fan Yang, and Mao Yang. 2025.
\newblock Mutual reasoning makes smaller {LLM}s stronger problem-solver.
\newblock In \emph{The Thirteenth International Conference on Learning Representations}.

\bibitem[{Qing et~al.(2024)Qing, Gao, Zhou, Diao, Yang, and Vosoughi}]{qing-etal-2024-alphalora}
Peijun Qing, Chongyang Gao, Yefan Zhou, Xingjian Diao, Yaoqing Yang, and Soroush Vosoughi. 2024.
\newblock {A}lpha{L}o{RA}: Assigning {L}o{RA} experts based on layer training quality.
\newblock In \emph{Proceedings of the 2024 Conference on Empirical Methods in Natural Language Processing}.

\bibitem[{Rajbhandari et~al.(2020)Rajbhandari, Rasley, Ruwase, and He}]{10.5555/3433701.3433727}
Samyam Rajbhandari, Jeff Rasley, Olatunji Ruwase, and Yuxiong He. 2020.
\newblock Zero: memory optimizations toward training trillion parameter models.
\newblock In \emph{Proceedings of the International Conference for High Performance Computing, Networking, Storage and Analysis}.

\bibitem[{Seddik et~al.(2024)Seddik, Chen, Hayou, Youssef, and Debbah}]{seddik2024bad}
Mohamed El~Amine Seddik, Suei-Wen Chen, Soufiane Hayou, Pierre Youssef, and Merouane Debbah. 2024.
\newblock How bad is training on synthetic data? a statistical analysis of language model collapse.
\newblock \emph{arXiv preprint arXiv:2404.05090}.

\bibitem[{Shi et~al.(2025)Shi, Fu, Yuan, Yu, You, Li, Dong, Kautz, Molchanov, and Lin}]{shilacache}
Dachuan Shi, Yonggan Fu, Xiangchi Yuan, Zhongzhi Yu, Haoran You, Sixu Li, Xin Dong, Jan Kautz, Pavlo Molchanov, and Yingyan~Celine Lin. 2025.
\newblock Lacache: Ladder-shaped kv caching for efficient long-context modeling of large language models.
\newblock In \emph{Forty-second International Conference on Machine Learning}.

\bibitem[{Shi et~al.(2023)Shi, Tao, Rao, Yang, Yuan, and Wang}]{shi2023crossget}
Dachuan Shi, Chaofan Tao, Anyi Rao, Zhendong Yang, Chun Yuan, and Jiaqi Wang. 2023.
\newblock Crossget: Cross-guided ensemble of tokens for accelerating vision-language transformers.
\newblock \emph{arXiv preprint arXiv:2305.17455}.

\bibitem[{Shumailov et~al.(2023)Shumailov, Shumaylov, Zhao, Gal, Papernot, and Anderson}]{shumailov2023curse}
Ilia Shumailov, Zakhar Shumaylov, Yiren Zhao, Yarin Gal, Nicolas Papernot, and Ross Anderson. 2023.
\newblock The curse of recursion: Training on generated data makes models forget.
\newblock \emph{arXiv preprint arXiv:2305.17493}.

\bibitem[{Silver et~al.(2018)Silver, Hubert, Schrittwieser, Antonoglou, Lai, Guez, Lanctot, Sifre, Kumaran, Graepel et~al.}]{silver2018general}
David Silver, Thomas Hubert, Julian Schrittwieser, Ioannis Antonoglou, Matthew Lai, Arthur Guez, Marc Lanctot, Laurent Sifre, Dharshan Kumaran, Thore Graepel, et~al. 2018.
\newblock A general reinforcement learning algorithm that masters chess, shogi, and go through self-play.
\newblock \emph{Science}, 362(6419):1140--1144.

\bibitem[{Singh et~al.(2023)Singh, Co-Reyes, Agarwal, Anand, Patil, Garcia, Liu, Harrison, Lee, Xu et~al.}]{singh2023beyond}
Avi Singh, John~D Co-Reyes, Rishabh Agarwal, Ankesh Anand, Piyush Patil, Xavier Garcia, Peter~J Liu, James Harrison, Jaehoon Lee, Kelvin Xu, et~al. 2023.
\newblock Beyond human data: Scaling self-training for problem-solving with language models.
\newblock \emph{arXiv preprint arXiv:2312.06585}.

\bibitem[{Snell et~al.(2025)Snell, Lee, Xu, and Kumar}]{snell2024scaling}
Charlie~Victor Snell, Jaehoon Lee, Kelvin Xu, and Aviral Kumar. 2025.
\newblock Scaling {LLM} test-time compute optimally can be more effective than scaling parameters for reasoning.
\newblock In \emph{The Thirteenth International Conference on Learning Representations}.

\bibitem[{Sutskever(2024)}]{ilya2024sequence}
Ilya Sutskever. 2024.
\newblock Sequence to sequence learning with neural networks: what a decade.
\newblock \emph{Keynote at NeurIPS}.

\bibitem[{Touvron et~al.(2023)Touvron, Martin, Stone, Albert, Almahairi, Babaei, Bashlykov, Batra, Bhargava, Bhosale et~al.}]{touvron2023llama}
Hugo Touvron, Louis Martin, Kevin Stone, Peter Albert, Amjad Almahairi, Yasmine Babaei, Nikolay Bashlykov, Soumya Batra, Prajjwal Bhargava, Shruti Bhosale, et~al. 2023.
\newblock Llama 2: Open foundation and fine-tuned chat models.
\newblock \emph{arXiv preprint arXiv:2307.09288}.

\bibitem[{Wang et~al.(2023)Wang, Le, Gotmare, Bui, Li, and Hoi}]{wang2023codet5+}
Yue Wang, Hung Le, Akhilesh~Deepak Gotmare, Nghi~DQ Bui, Junnan Li, and Steven~CH Hoi. 2023.
\newblock Codet5+: Open code large language models for code understanding and generation.
\newblock \emph{arXiv preprint arXiv:2305.07922}.

\bibitem[{Wei et~al.(2022)Wei, Wang, Schuurmans, Bosma, Xia, Chi, Le, Zhou et~al.}]{wei2022chain}
Jason Wei, Xuezhi Wang, Dale Schuurmans, Maarten Bosma, Fei Xia, Ed~Chi, Quoc~V Le, Denny Zhou, et~al. 2022.
\newblock Chain-of-thought prompting elicits reasoning in large language models.
\newblock \emph{Advances in neural information processing systems}.

\bibitem[{Wu et~al.(2024)Wu, Yuan, Golovneva, Xu, Tian, Jiao, Weston, and Sukhbaatar}]{wu2024meta}
Tianhao Wu, Weizhe Yuan, Olga Golovneva, Jing Xu, Yuandong Tian, Jiantao Jiao, Jason Weston, and Sainbayar Sukhbaatar. 2024.
\newblock Meta-rewarding language models: Self-improving alignment with llm-as-a-meta-judge.
\newblock \emph{arXiv preprint arXiv:2407.19594}.

\bibitem[{Wu et~al.(2025)Wu, Xu, Gao, Diao, Li, Salas, and Gui}]{wu2025assessing}
Weiyi Wu, Xinwen Xu, Chongyang Gao, Xingjian Diao, Siting Li, Lucas~A Salas, and Jiang Gui. 2025.
\newblock Assessing and mitigating medical knowledge drift and conflicts in large language models.
\newblock \emph{arXiv preprint arXiv:2505.07968}.

\bibitem[{Yang and et~al.(2024)}]{qwen2.5}
An~Yang and et~al. 2024.
\newblock Qwen2.5 technical report.
\newblock \emph{arXiv preprint arXiv:2412.15115}.

\bibitem[{Yao et~al.(2024)Yao, Yu, Zhao, Shafran, Griffiths, Cao, and Narasimhan}]{yao2024tree}
Shunyu Yao, Dian Yu, Jeffrey Zhao, Izhak Shafran, Tom Griffiths, Yuan Cao, and Karthik Narasimhan. 2024.
\newblock Tree of thoughts: Deliberate problem solving with large language models.
\newblock \emph{Advances in Neural Information Processing Systems}.

\bibitem[{Ye et~al.(2025)Ye, Xia, Fu, Dong, Hong, Yuan, Diao, Kautz, Molchanov, and Lin}]{ye2025longmamba}
Zhifan Ye, Kejing Xia, Yonggan Fu, Xin Dong, Jihoon Hong, Xiangchi Yuan, Shizhe Diao, Jan Kautz, Pavlo Molchanov, and Yingyan~Celine Lin. 2025.
\newblock Longmamba: Enhancing mamba's long context capabilities via training-free receptive field enlargement.
\newblock \emph{The Thirteenth International Conference on Learning Representations}.

\bibitem[{Yu et~al.(2024{\natexlab{a}})Yu, Yu, Yu, Huang, and Li}]{yu2024language}
Le~Yu, Bowen Yu, Haiyang Yu, Fei Huang, and Yongbin Li. 2024{\natexlab{a}}.
\newblock Language models are super mario: Absorbing abilities from homologous models as a free lunch.
\newblock In \emph{International Conference on Machine Learning}.

\bibitem[{Yu et~al.(2024{\natexlab{b}})Yu, Jiang, Shi, YU, Liu, Zhang, Kwok, Li, Weller, and Liu}]{yu2024metamath}
Longhui Yu, Weisen Jiang, Han Shi, Jincheng YU, Zhengying Liu, Yu~Zhang, James Kwok, Zhenguo Li, Adrian Weller, and Weiyang Liu. 2024{\natexlab{b}}.
\newblock Metamath: Bootstrap your own mathematical questions for large language models.
\newblock In \emph{ICLR}.

\bibitem[{Yuan et~al.(2024)Yuan, Pang, Cho, Sukhbaatar, Xu, and Weston}]{yuan2024self}
Weizhe Yuan, Richard~Yuanzhe Pang, Kyunghyun Cho, Sainbayar Sukhbaatar, Jing Xu, and Jason Weston. 2024.
\newblock Self-rewarding language models.
\newblock \emph{arXiv preprint arXiv:2401.10020}.

\bibitem[{Yuan et~al.(2023)Yuan, Yuan, Li, Dong, Lu, Tan, Zhou, and Zhou}]{yuan2023scaling}
Zheng Yuan, Hongyi Yuan, Chengpeng Li, Guanting Dong, Keming Lu, Chuanqi Tan, Chang Zhou, and Jingren Zhou. 2023.
\newblock Scaling relationship on learning mathematical reasoning with large language models.
\newblock \emph{arXiv preprint arXiv:2308.01825}.

\bibitem[{Zelikman et~al.(2022)Zelikman, Wu, Mu, and Goodman}]{zelikman2022star}
Eric Zelikman, Yuhuai Wu, Jesse Mu, and Noah Goodman. 2022.
\newblock Star: Bootstrapping reasoning with reasoning.
\newblock \emph{Advances in Neural Information Processing Systems}.

\bibitem[{Zhang et~al.(2024)Zhang, Jian, Ouyang, and Vosoughi}]{zhang2024working}
Chunhui Zhang, Yiren Jian, Zhongyu Ouyang, and Soroush Vosoughi. 2024.
\newblock Working memory identifies reasoning limits in language models.
\newblock In \emph{Proceedings of the 2024 Conference on Empirical Methods in Natural Language Processing}.

\bibitem[{Zhang et~al.(2025{\natexlab{a}})Zhang, Ouyang, Lee, Agarwal, Houlihan, Vosoughi, and Lo}]{zhang2025overcoming}
Chunhui Zhang, Zhongyu Ouyang, Kwonjoon Lee, Nakul Agarwal, Sean~Dae Houlihan, Soroush Vosoughi, and Shao-Yuan Lo. 2025{\natexlab{a}}.
\newblock \href {https://openreview.net/forum?id=2dz6psiiA0} {Overcoming multi-step complexity in multimodal theory-of-mind reasoning: A scalable bayesian planner}.
\newblock In \emph{Forty-second International Conference on Machine Learning}.

\bibitem[{Zhang et~al.(2025{\natexlab{b}})Zhang, Ouyang, Yuan, Vosoughi et~al.}]{zhang2025growing}
Chunhui Zhang, Zhongyu Ouyang, Xiangchi Yuan, Soroush Vosoughi, et~al. 2025{\natexlab{b}}.
\newblock Growing through experience: Scaling episodic grounding in language models.
\newblock \emph{The 63rd Annual Meeting of the Association for Computational Linguistics}.

\bibitem[{Zheng et~al.(2024)Zheng, Mishra, Chen, Cheng, Chi, Le, and Zhou}]{zheng2023take}
Huaixiu~Steven Zheng, Swaroop Mishra, Xinyun Chen, Heng-Tze Cheng, Ed~H. Chi, Quoc~V Le, and Denny Zhou. 2024.
\newblock Take a step back: Evoking reasoning via abstraction in large language models.
\newblock In \emph{The Twelfth International Conference on Learning Representations}.

\bibitem[{Zhong et~al.(2024)Zhong, Cui, Guo, Liang, Lu, Wang, Saied, Chen, and Duan}]{zhong-etal-2024-agieval}
Wanjun Zhong, Ruixiang Cui, Yiduo Guo, Yaobo Liang, Shuai Lu, Yanlin Wang, Amin Saied, Weizhu Chen, and Nan Duan. 2024.
\newblock {AGIE}val: A human-centric benchmark for evaluating foundation models.
\newblock In \emph{Findings of the Association for Computational Linguistics: NAACL 2024}.

\bibitem[{Zhou and Srikumar(2021)}]{zhou2021closer}
Yichu Zhou and Vivek Srikumar. 2021.
\newblock A closer look at how fine-tuning changes bert.
\newblock \emph{arXiv preprint arXiv:2106.14282}.

\end{thebibliography}
\clearpage
\appendix


\section{Training and Implementation Details}
\subsection{Chain of Thought Prompting for Data Synthesis}
\label{app:cot}

We use chain-of-thought prompting \cite{wei2022chain} to generate answers. For MATH and GSM8K datasets, we both give 10 examples in the instructions for in-context generation. The prompting examples are given in Table~\ref{tab:cot_gsm8k}. We generate 3 candidate answers for GSM8K and 6 candidate answers for MATH to have comparable numbers of right answers.
\subsection{Training Details}
\label{app:training_detail}
We use NVIDIA RTX 8 $\times$ A6000 to train the model with DeepSpeed \cite{10.5555/3433701.3433727} distributed training framework.
The number of training epoch is 3 and per device training batch size is 4. The gradient accumulation steps are set to 4 and the learning rate is 2e-5. The warm-up rate is 0.03. We use mixed precision training with bf16.
We use DeepSpeed to distribute supervised fine-tuning model with ZeRO3, which partitions all three model states. We also use the vLLM library \cite{kwon2023efficient} to generate synthetic reasoning data with sampling temperature \{0.2, 0.4, 0.6\} to balance the diversity and accuracy of generated answers. Note that we use all models, data, and training tools solely for research purpose, which are consistent with their intended use.

The Model merging parameter in Section~\ref{sec:motivation} is set to 0.5 to balance the base model and self-improved model. We use the setting in Section~\ref{sec:distill} to do the parameter analysis for $\alpha$. Table~\ref{tab:alpha} shows that $\alpha=0.5$ can achieve a good balance between ID and OOD performance.

\begin{table}[h]
\resizebox{0.48\textwidth}{!}{%
\begin{tabular}{lccccccc}
\toprule
$\alpha$        & 0.1  & 0.3  & 0.4           & 0.5           & 0.6  & 0.7  & 0.9           \\ \midrule
GSM8K     & 3.7  & 24.5 & 32.4          & 38.8          & 40.4 & 43.3 & \textbf{48.3} \\
MATH      & 3.7  & 8.4  & 10.8          & \textbf{12.5} & 12.5 & 11.8 & 10.3          \\
SAT\_Math & 25.3 & 27.4 & 27.7          & \textbf{28.1} & 26.7 & 24.8 & 18.8          \\
MAWPS     & 64.4 & 69.0 & \textbf{76.8} & 76.6          & 69.3 & 64.3 & 57.2          \\ \bottomrule
\end{tabular}}
\caption{The parameter analysis for $\alpha$.}
\label{tab:alpha}
\end{table}

\begin{table*}[htb]
\label{tab:app_ood}
\begin{tabular}{lcccccccc}
\toprule
Model                           & Datasets   & Base          & SFT1          & Merge1 & SFT2 & Merge2        & SFT3          & Merge3        \\ \midrule
\multirow{4}{*}{Qwen2.5-0.5B-I} & SVAMP      & 7.3           & 35.8          & 5.9    & 21.6 & 1.3           & \textbf{40.1} & 9.8           \\
                                & ASDiv      & 8.7          & \textbf{51.4} & 3.7    & 30.7 & 2.8           & 46.7          & 17.6          \\
                                & MathQA     & \textbf{37.9} & 29.5          & 38.2   & 25.7 & 35.4          & 19.9          & 33.8          \\
                                & MMLU\_stem & 34.2          & 34.6          & 38.4   & 34.3 & \textbf{37.1} & 27.9          & 36.1          \\ \midrule
\multirow{4}{*}{Qwen2.5-1.5B-I} & svamp      & \textbf{77.7} & 59            & 69.2   & 58.6 & 58.2          & 60.2          & 64.7          \\
                                & asdiv      & \textbf{82.8} & 72.5          & 76.4   & 64.8 & 59.6          & 70.8          & 73.4          \\
                                & MathQA     & \textbf{62.5} & 24.9          & 57.3   & 33.4 & 54.1          & 12.8          & 53.4          \\
                                & MMLU\_stem & 53.6          & 40.1          & 52.6   & 47.9 & 53.4          & 41.7          & \textbf{54.5} \\ \midrule
\multirow{2}{*}{Llama2-7B}      & svamp      & 39.6          & 30.1          & 38.0     & 35.1 & 39.0            & 33.5          & 38.5          \\
                                & asdiv      & 51.9          & 42.9          & 51.2   & 46.7 & 52.3          & 41.4          & \textbf{52.7} \\ \bottomrule
\end{tabular}
\caption{ OOD performance on additional reasoning datasets.}
\label{tab:ood_app}
\end{table*}

\begin{table*}[h]
\centering
\begin{tabular}{lcccc}
\toprule
Datasets                          & GSM8K         & MATH          & SAT\_Math     & MAWPS         \\ \midrule
Vanilla SFT                       & 58.5          & 32.5          & 50.0          & 85.9          \\
Gradient-decay ($\gamma$=0.9)     & 59.2          & 32.8          & 53.8          & 84.2          \\
Gradient-clipping (max\_norm=2.0) & 58.7          & 31.7          & 52.3          & 84.7          \\
Weight-masking (TopP=0.3)         & 60.2          & \textbf{34.5} & 56.2          & 87.0          \\
IMM                               & \textbf{69.3} & 34.0          & \textbf{68.8} & \textbf{89.4} \\ \bottomrule
\end{tabular}
\caption{Qwen2.5-1.5B-Instruct performance compared with SFT interventions in the first iteration.}
\label{tab:interventions}
\end{table*}

\section{Additional Experiments and Analysis}

\label{app:exp}

\subsection{Superficial Reasoning Finetuning Exists When Real Data Is Limited}
\label{app:real_data}
\begin{figure}[h]  
    \centering
    \includegraphics[width=0.5\textwidth]{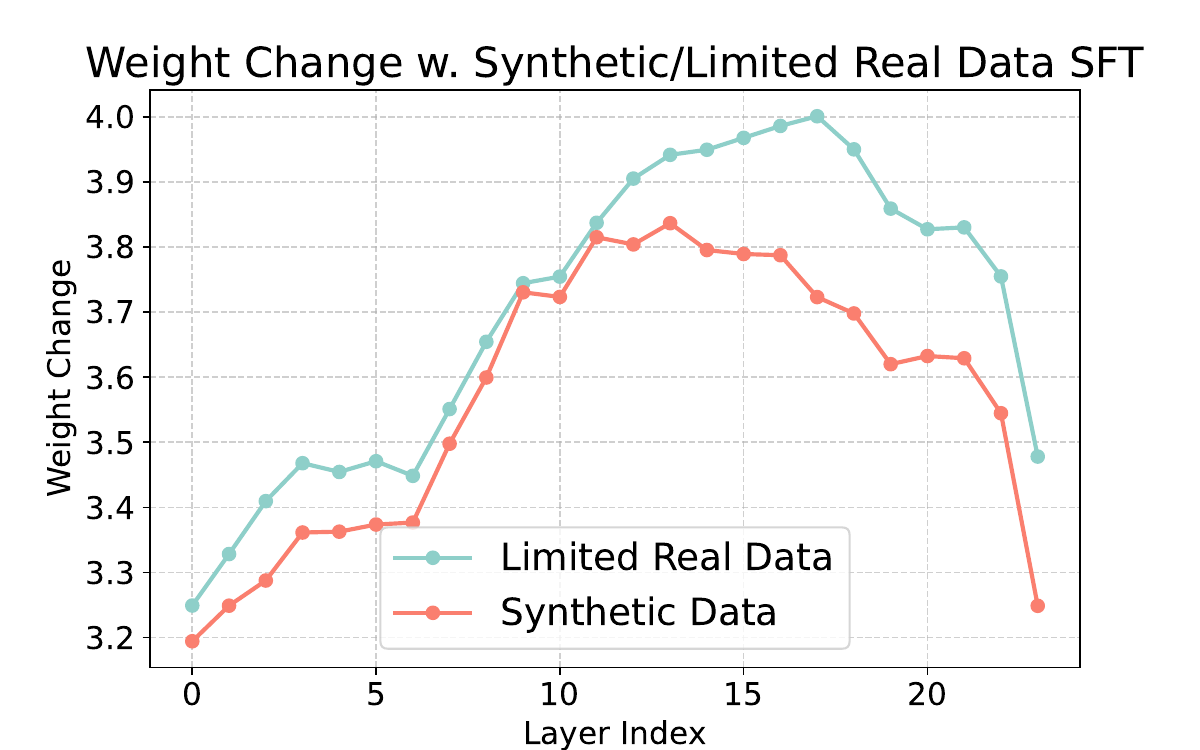}
    \caption{The weight change over layers for (i) Fintuning Qwen2.5-1.5B with synthetic MATH \cite{hendrycks2021measuring} dataset data and limited training data (7.5k real MATH training data)}
    \label{fig:real_syn_both}
\end{figure}

We also find that even using real but limited data, Superficial Reasoning Synthetic Finetuning still exists. As Figure \ref{fig:real_syn_both} shows, the middle layers change most compared with the early and late layers, while Figure \ref{layer_importance} already shows that early and late layers are more important for reasoning. However, utilizing real data prevents the model from overfitting itself by using self-generated data. This is also verified by Figure \ref{fig:real_syn_both}: the model's reasoning layer (early and late layers) changed more (learn more reasoning capability) when training with real data, the reasoning-trivial layers (middle layers)'s weight change is close to middle layers when training with synthetic data. 

\subsection{Layer Importance}
\begin{figure}[h]  
    \centering
    \includegraphics[width=0.5\textwidth]{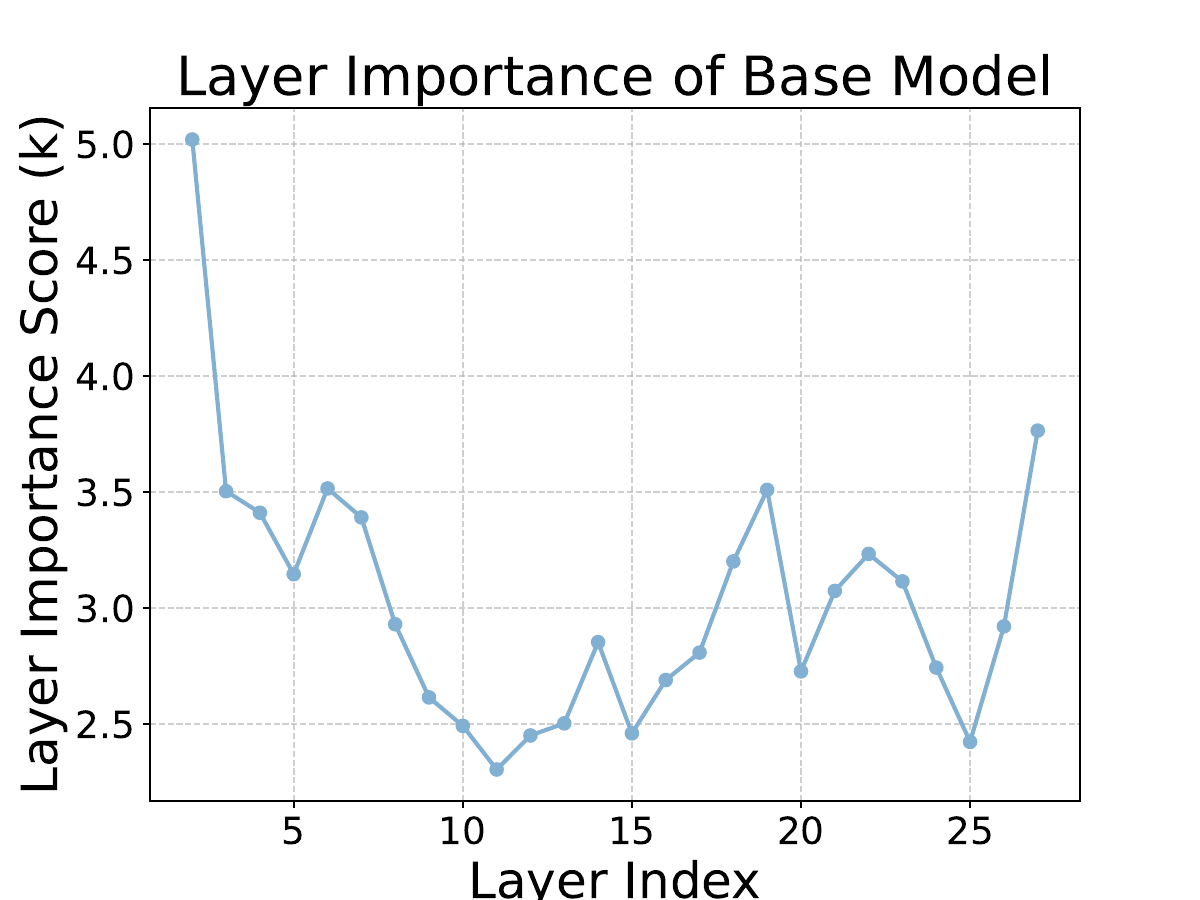}
    \caption{The layer importance score for Qwen2.5-1.5B base model on reasoning dataset MATH.}
    \label{fig:import_base_math}
\end{figure}

\begin{figure}[h]  
    \centering
    \includegraphics[width=0.5\textwidth]{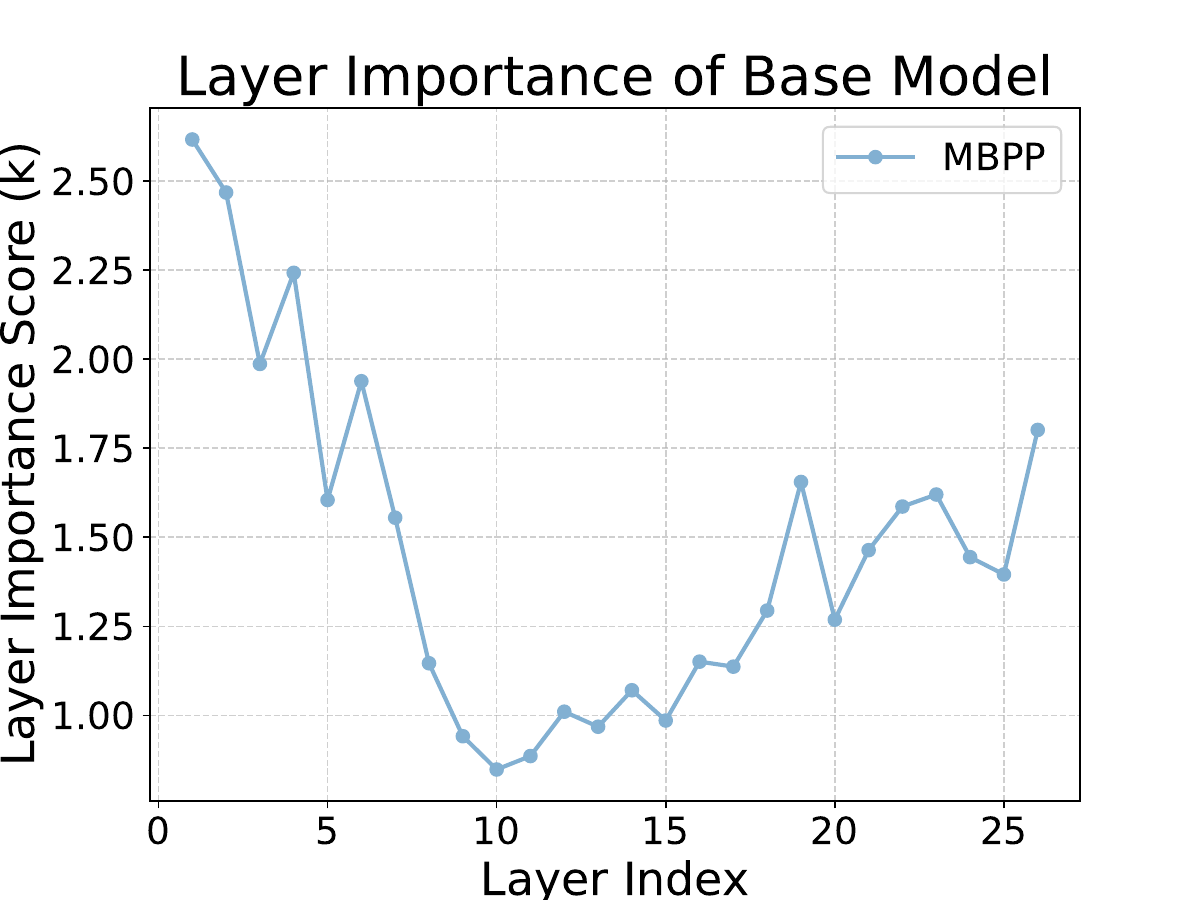}
    \caption{The layer importance score for Qwen2.5-1.5B base model on reasoning dataset MBPP.}
    \label{fig:import_base_code}
\end{figure}
\label{app:exp_layer_i}
Here we provide the additional experiment results for evaluating the layer importance for Qwen2.5-1.5B base model on reasoning datasets MATH. Similar to stronger reasoning model Qwen2.5-1.5B-Math, the importance layer for reasoning is early and late layers, as demonstrated in Figure~\ref{fig:import_base_math}. We also observed similar behavior in other complex reasoning task code generation MBPP, as demonstrated in Figure~\ref{fig:import_base_code}.

\subsection{OOD Performance}
\label{app:ood_exp}
We also provide OOD performance on additional datasets SVAMP \cite{patel-etal-2021-nlp}, ASDiv \cite{miao-etal-2020-diverse}, MathQA \cite{amini2019mathqa} and MMLU-stem \citet{hendrycks2020measuring}. IMM keeps the OOD reasoning capability as shown in Table~\ref{tab:ood_app}.

\subsection{Comparison with SFT Interventions}
\label{app:sft_inter}
We provide experimental results to compare these alternative interventions with IMM. As shown in Table~\ref{tab:interventions}, these methods generally do not outperform IMM, and in some cases are even outperformed by vanilla SFT.
Compared with these interventions during SFT, IMM not only mitigate the overfitting reasoning finetuning, but also improve generalized reasoning capability through ensemble model merging. Our method is orthogonal to interventions for SFT, and provides a simple yet effective method to solve superficial self-improved reasoners phenomenon identified by this research. 

\subsection{Importance-based Weight Merge}
\label{app:iimm}

\begin{table}[h]
\centering
\resizebox{\linewidth}{!}{%
\begin{tabular}{lcccc}
\toprule
\multicolumn{1}{c}{Datasets} & GSM8K         & MATH          & SAT\_Math     & \multicolumn{1}{c}{MAWPS} \\ \midrule
I-IMM                        & \textbf{44.2} & \textbf{27.5} & 49.3          & 2.1                       \\
IMM                          & \textbf{44.2} & 27.4          & \textbf{56.2} & \textbf{3.4}              \\ \bottomrule
\end{tabular}
}
\caption{Comparison of IIMM and IMM across ID and OOD datasets.}
\label{tab:exp-iimm}
\end{table}

We also experimented with weighting the merge ratio $\alpha$ per layer using the importance score $I$ defined in Eq. (3). As shown in Table~\ref{tab:exp-iimm}, this approach occasionally improves in-domain (ID) performance but often performs worse on out-of-domain (OOD) datasets. 
We hypothesize that this is because weighting the merging process based on ID-specific importance scores leads to overfitting to the ID data, thereby sacrificing the model’s generalized reasoning capabilities on OOD tasks. Additionally, imbalanced merging rates across layers may introduce instability: when different layers are merged to varying degrees, the model can become internally inconsistent. In an extreme case, if some layers remain largely as base model layers while others are heavily adapted via SFT, this imbalance can degrade performance, as the layers are no longer "on the same page". We consider our merging method as a new way to increase the generalization of supervised learning, like other methods such as regularization~\cite{jin2024learning} and meta-tuning~\cite{guo-etal-2024-meta}.

\subsection{Analysis on Unexpected Behavior}
\label{app:unexpected_behavior}
\begin{table}[]
\centering
\begin{tabular}{lccc}
\toprule
Model        & Base          & SFT           & Merge         \\ \midrule
Qwen2.5-0.5B & 12.8          & \textbf{32.9} & {\ul 23.4}    \\
Qwen2.5-1.5B & \textbf{90.0} & {\ul 72.8}    & 24.5          \\
Llama-2-7B   & {\ul 64.1}    & 52.6          & \textbf{65.3} \\ \bottomrule
\end{tabular}
\caption{Model performances on MAWPS dataset. The best performances are \textbf{bolded}, and the runner-up performances are {\ul underlined}.}
\label{tab:mawps}
\end{table}

\begin{table}[]
\centering
\begin{tabular}{lccc}
\toprule
Dataset    & Base          & SFT        & Merge         \\ \midrule
SAT\_Math  & {\ul 75.0}    & {\ul 75.0} & \textbf{87.5} \\
MAWPS      & \textbf{90.0} & {\ul 72.8} & 24.5          \\
MathQA     & \textbf{62.5} & 55.5       & {\ul 62.0}    \\
MMLU\_stem & {\ul 53.6}    & 54.5       & \textbf{57.6} \\
SVAMP      & \textbf{77.7} & 54.1       & {\ul 61.2}    \\ \bottomrule
\end{tabular}
\caption{Qwen2.5-1.5B-Instruct performance on external OOD datasets. The best performances are \textbf{bolded}, and the runner-up performances are {\ul underlined}.}
\label{tab:1.5bood}
\end{table}

OOD performance drops for Qwen2.5-1.5B on MAWPS dataset, and here we conduct more experiments to analyze this behavior. We found that (Table~\ref{tab:mawps}) small models (e.g., 0.5B and 1.5B) only suffer significant performance degradation on the MAWPS dataset after model merging. In contrast, larger models (e.g., 7B) achieve the best performance on MAWPS, benefiting more from IMM. Despite this drop on MAWPS, smaller models still show performance improvements on other OOD datasets. For instance, Table~\ref{tab:1.5bood} shows that the 1.5B model outperforms both the Base and SFT versions on 5 OOD datasets. Therefore, we attribute the performance degradation on MAWPS primarily to two factors: (1) potential distributional differences in MAWPS compared to other datasets, and (2) the limited parameter capacity of small models, which may lack sufficient redundancy to support robust merging without trade-offs.

\subsection{Additional Test-time Computing Results}
\label{app:test_time}
\begin{figure}[]

    \centering
    \begin{minipage}[t]{0.24\textwidth}
        \centering
        \includegraphics[width=\linewidth]{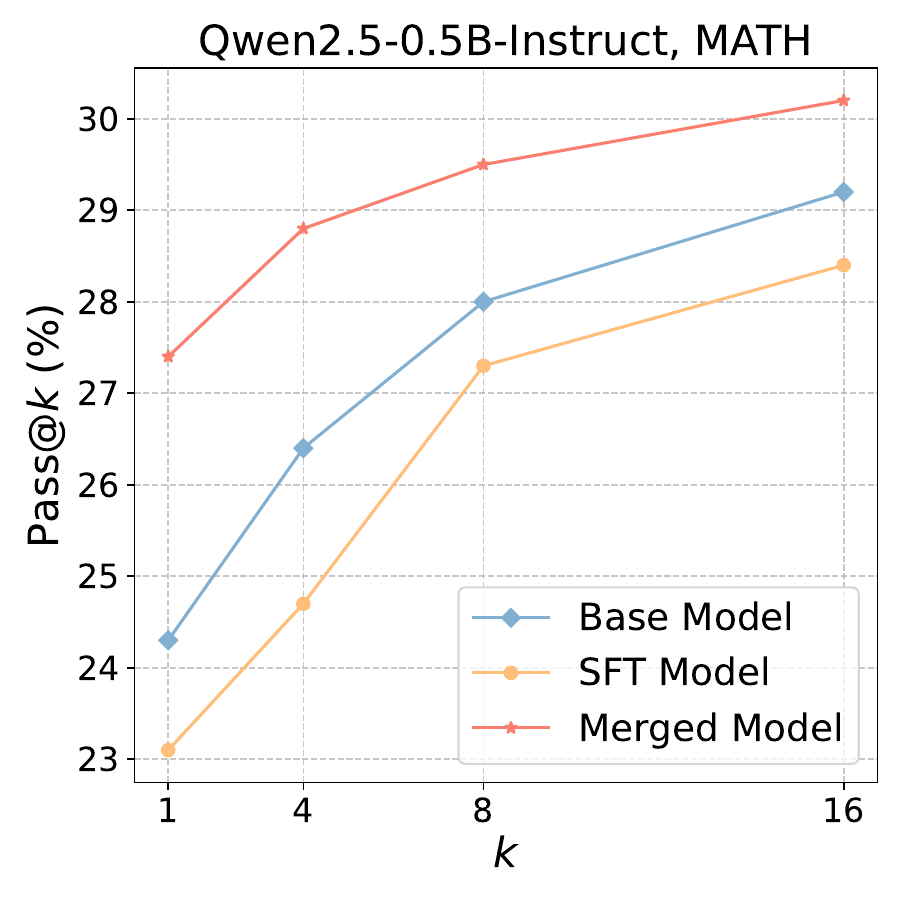}
    \end{minipage}%
    \begin{minipage}[t]{0.24\textwidth}
        \centering
        \includegraphics[width=\linewidth]{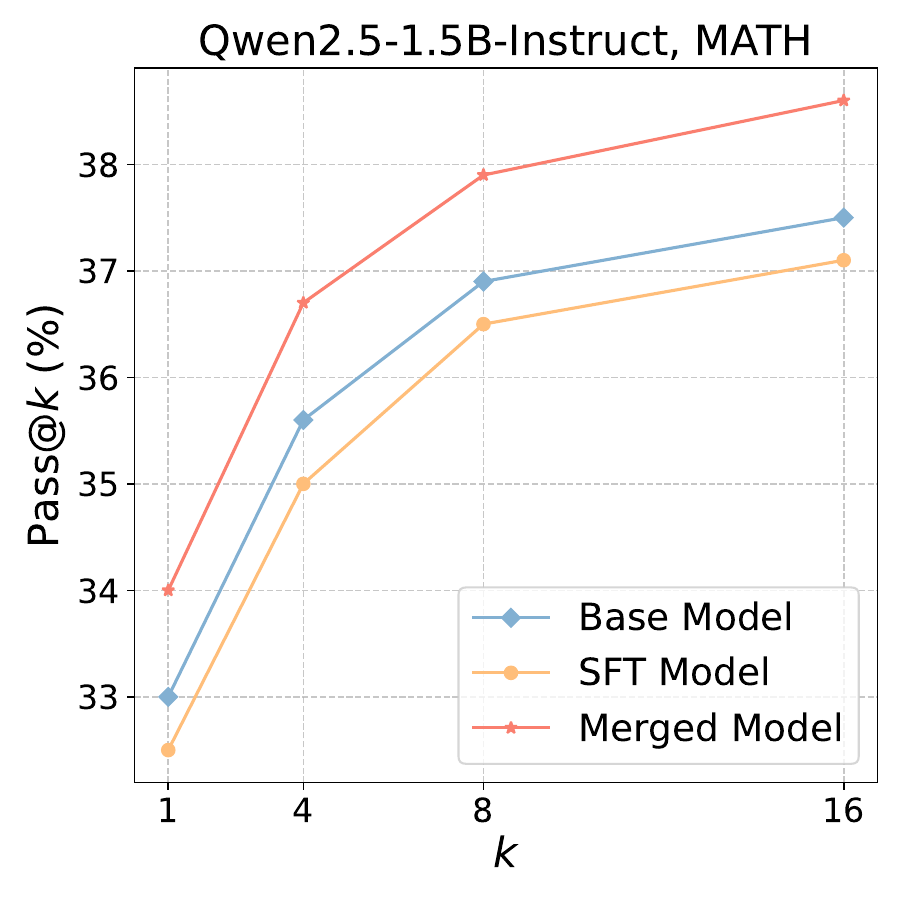}
    \end{minipage}%

    \caption{ID performance with different k for scaling up test-time-computing Pass@k on MATH.}
    \label{fig:pass_math}
    \vspace{-5pt}
\end{figure}

We evaluate our method by generating multiple answers and measuring pass@$k$ accuracy for MATH dataset. As shown in Figure \ref{fig:pass_math}, our method consistently improves performance as k increases and outperforms both the base models and the SFT models.

\subsection{IMM with the Recent Model}
\label{app:llama3}
Table~\ref{tab:llama3} shows that for Llama3-8B model, IMM improves the ID performance and keeps comparable OOD performance, while vanilla SFT suffers from model collapse in ID datasets and severe degradation on OOD datasets.

\begin{figure}[h]  
    \centering
    \includegraphics[width=0.5\textwidth]{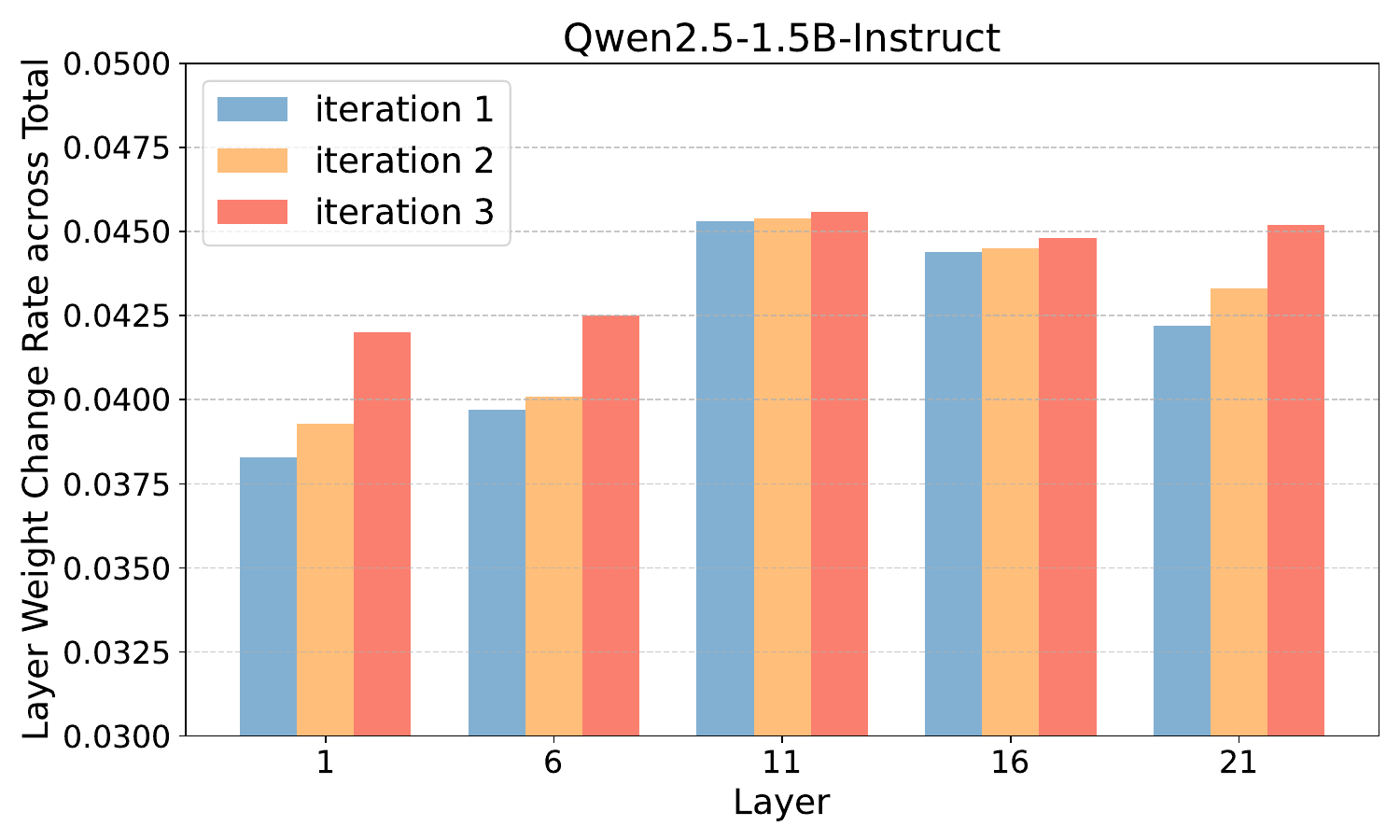}
    \caption{The percentage of the weight change over layers for finetuning Qwen2.5-1.5B in different iterations.}
    \label{fig:change_percentage}
\end{figure}

\subsection{Weight Change for Different Iterations}
\label{app:weight_change_iterations}
After the first model weight merge, the parameter updates for the layers critical for reasoning still remain minimal, as illustrated in Figure~\ref{weight_change}. However, we continue to analyze the weight change across different layers and find that, although IMM uses an average merge rate across different layers, it improves the model's generalized reasoning capability, which makes the weight of reasoning-critical layers change more in the next iterations. Figure~\ref{fig:change_percentage} shows that, in the next iterations, the reasoning-critical layers (early and late layers) change more weight change compared with the reasoning-trivial layers (middle layers), indicating the model learns the generalized reasoning capability after IMM. Also, although IMM uses a uniform merge rate $\alpha$ across all layers, the absolute weight change difference between reasoning-critical layers and reasoning-trivial layers becomes smaller compared with SFT. This small difference accumulates over the course of the iterative self-improvement process. As a result, IMM achieves a relatively more balanced distribution of weight changes across layers compared to vanilla self-improvement and other baselines, where middle layers undergo disproportionately larger updates than early and late layers.
IMM model therefore brings better generalized reasoning capability.

\begin{table}[]
\resizebox{\linewidth}{!}{%
\begin{tabular}{lcccc}
\toprule
\multicolumn{1}{c}{Datasets} & GSM8K         & MATH          & SAT\_Math           & MAWPS         \\ \midrule
Base                         & {\ul 55.1}    & 16.1          & {\ul \textbf{53.1}} & \textbf{90.8} \\
SFT                          & 53.4          & {\ul 17.2}    & 35.2                & 80.1          \\
IMM                          & \textbf{61.2} & \textbf{19.5} & 52.8                & {\ul 89.5}    \\ \bottomrule
\end{tabular}}

\caption{Llama3-8B performance for the first self-improvement iteration. The best performances are \textbf{bolded}, and the runner-up performances are {\ul underlined}.}
\label{tab:llama3}
\end{table}

\subsection{Complexity}
\label{app:complexity}
Let $n$ be the number of model parameters, $T$ be the number of IMM iterations, $F(n)$ be the cost of one SFT training session. We calculate the complexity for IMM in the Table~\ref{tab:imm-complexity}. The overall complexity is $\mathcal{O}(T \cdot F(n))$. Since fine-tuning dominates, especially for large models, the primary bottleneck is still the repeated SFT stages. Therefore, IMM introduce linear complexity on $n$, which can be overlooked compared with $\mathcal{O}(F(n))$, ensuring the scalability.

\section{Additional Discussion and Clarification}

\subsection{A Bitter Lesson: Not All LLMs Can Self-improve}
During our experiment, we also find that not all the LLMs can self-improve on reasoning tasks. If LLM's performance decreases after SFT, then our method may not let the merged model have a better performance compared with the original model and the model after SFT.  This usually happens when the original model already has a good performance (reasoning ability), and learned reasoning ability can't offset the generalization loss.

\subsection{Why Importance-Weight Change Mismatch Happens?}
\label{app:why_mismatch}
We conclude two possible contributing factors to this observation:
(i) Characteristics of SFT on Pre-trained LMs: Prior studies \cite{merchant-etal-2020-happens,mosbach-etal-2020-interplay,zhou2021closer} have shown that during SFT, the early and late layers of pre-trained language models tend to undergo minimal changes. In particular, the late layers often preserve their original representations, suggesting a structural bias of SFT toward updating the middle layers.
(ii) Inhibitory Effect of Self-improvement on Reasoning-critical Layers: As shown in Figure~\ref{fig:real_syn_both}, models fine-tuned on real data exhibit more weight change in reasoning-critical layers (early and late layers) compared to those fine-tuned on self-synthesized data. In contrast, the middle layers show comparable levels of weight change in both settings. This indicates that the self-improvement process inherently inhibits updates to reasoning-critical layers, leading to disproportionate changes in the middle layers.

We further explain why middle layers contribute less to complex reasoning tasks. Prior work \cite{li2024understanding} shows that weaker, implicit reasoning signals tend to surface in the middle layers, whereas stronger, explicit reasoning—such as chain-of-thought reasoning—emerges primarily in the late (and occasionally early) layers. In our study, to solve complex reasoning tasks model generated long CoT reasoning path, which depends on late layers

In summary, superficial self-improvement leads to overfitting on middle-layer representations where weaker, implicit reasoning resides, due to both the inherent bias of SFT and self-generated data. In contrast, reasoning-critical layers, responsible for explicit CoT reasoning, remain largely unchanged, limiting the model’s ability to improve on more complex reasoning tasks. Actually, it's very common that different layer has different behaviors for tasks~\cite{qing-etal-2024-alphalora}, and our work emphasizes it in reasoning.

\begin{table}[]
\centering
\resizebox{\linewidth}{!}{%
\begin{tabular}{lc}
\toprule
Operation & Complexity \\
\midrule
SFT & $\mathcal{O}(F(n))$ \\
Compute $\delta^t$ & $\mathcal{O}(n)$ \\
Masking, scaling & $\mathcal{O}(n)$ \\
Merge update & $\mathcal{O}(n)$ \\

Overall Complexity & $\mathcal{O}(T \cdot (F(n) + n)) \approx \mathcal{O}(T \cdot F(n))$ \\
\bottomrule
\end{tabular}}
\caption{Time complexity of IMM update steps.}
\label{tab:imm-complexity}
\end{table}

\subsection{The Connection to Catastrophic Forgetting}
\label{app:catastrophic}

Catastrophic forgetting is a related but distinct phenomenon compared to superficial self-improved reasoners. Specifically, catastrophic forgetting refers to the loss of previously acquired knowledge when deep learning models are trained on new data. This issue occurs because model parameters are optimized based on the most recent training data, causing earlier learned representations to be dramatically overwritten. 

While both catastrophic forgetting and superficial self-improved reasoners result in degraded performance due to further fine-tuning, their effects differ. After fine-tuning on new data, catastrophic forgetting results in a performance loss on previously learned tasks, whereas superficial self-improved reasoners result in diminished generalization capabilities on out-of-domain (OOD) tasks. 
This discrepancy arises because in catastrophic forgetting, fine-tuning on data for new tasks causes the model to lose knowledge from previous tasks. In contrast, superficial self-improved reasoners do not lead to forgetting too much past information but instead shift towards overfitting due to potentially biased knowledge, which may self-enhance along with the iteration of synthesizing new data and fine-tuning on it.

\subsection{The Definition for Layers}
\label{app:def_layers}
We do not provide a rigorous theoretical definition or external citation for the terms "reasoning-trivial layers" and "reasoning-trivial layers". In our paper, we adopt a relative and empirical definition: "reasoning-trivial layers" refer to the layers that exhibit lower importance scores in comparison to others, and "reasoning-trivial layers" refer to the layers that exhibit higher importance scores based on our layer-wise reasoning importance analysis. While not formally defined, this relative notion is sufficient for our purposes. It allows us to identify and analyze the mismatch between reasoning-critical layers (i.e., those with high importance scores) and the layers undergoing the most weight change during self-improvement. This mismatch is central to our discovery of the superficial self-improvement phenomenon.

\subsection{Why This Importance Score}
We would like to clarify that while the identification of key layers has been widely explored in prior work, such as in model analysis, pruning~\cite{liu2025modality}, and importance-based selection, our study does not aim to introduce a theoretical advancement in key layer selection itself. Rather, our contribution lies in uncovering a novel phenomenon: a mismatch between reasoning-critical layers and the layers experiencing the most weight change during self-improvement. We believe this observation offers a new perspective on how generalized reasoning capabilities may be hindered by superficial self-improvement. Building on this insight, we propose IMM as a method to mitigate this issue and improve the model’s generalization in reasoning tasks.

Compared to other popular evaluation such as gradient change, the metrics defined in Eq.~(\ref{eq:importance}) and Eq.~(\ref{eq:weight_change}) are more suitable for the type of analysis conducted in this work. Specifically, Eq.~(\ref{eq:importance}) directly measures "how much the parameters of a given layer have actually changed from the beginning to the end of training." This provides a clearer indication of how much information is retained or adjusted through the self-improvement process, which is more aligned with our goal of understanding where learning occurs across the model.
In contrast, gradient change is more appropriate for analyzing \textit{how quickly} or \textit{at which stage} the model learns during training. We appreciate the suggestion and agree that gradient analysis can provide complementary insights. We will include gradient tracking in manuscript to help monitor training stability and to identify potential issues such as exploding or vanishing gradients during self-improvement cycles.

\subsection{SFT Overfitting}
Overfitting is a common issue in supervised learning. Indeed, our work does not dispute or repeat this general principle. Rather, our contribution lies in highlighting a previously underexplored phenomenon: in particular, when applied to reasoning tasks, self-improvement via SFT tends to exacerbate generalization degradation more severely than standard SFT on curated or distilled data.

This distinction is central to our study. While overfitting is ubiquitous in supervised learning, our empirical results demonstrate that SFT using self-generated data amplifies this risk, leading to a sharper decline in reasoning generalization and more pronounced model collapse. This opinon is also supported by our interpretability analysis. To the best of our knowledge, this amplification effect of self-improvement has not been explicitly analyzed in prior work.
To support this claim, we conducted systematic experiments beyond the self-improvement setting in Section~\ref{sec:distill} and Appendix~\ref{app:real_data}.  We also demonstrate that even in general SFT settings like distillation, our proposed method still improves reasoning generalization, highlighting the broader applicability and potential of our approach.

\subsection{Model Scale and Overfitting}
While it is intuitive to assume that smaller models are more susceptible to overfitting, prior work~\cite{dohmatob2025strong} suggests that overfitting severity depends non-monotonically on model size. Specifically, they report that models below a certain size threshold may exhibit pronounced overfitting, whereas models above that threshold may also overfit due to reduced margin for improvement on certain tasks.

In our study (Section~\ref{sec:exp}), we scale model size to acknowledge this complexity and do not assert a universal trend. Our empirical results in Figure~\ref{fig:mian_id} and Figure~\ref{fig:mian_ood} show inconsistent overfitting behavior across different model sizes and datasets, making it difficult to isolate model size as the primary factor behind model collapse. In fact, our cross-model and cross-dataset comparison suggests an interesting trend: when a model already performs strongly on a dataset, further supervised fine-tuning on self-generated (and correct) samples may inadvertently trigger overfitting due to reduced room for meaningful generalization.

\subsection{How Other Methods Lead to  Superficial Reasoning Learning}
While existing approaches like data mixture \cite{shumailov2023curse}, data accumulation \cite{gerstgrasser2024is}, adding real data~\cite{dohmatob2025strong}, and data selection~\cite{guo-etal-2024-curious}, have indeed been effective in expanding the distribution or sampling diversity, their application has primarily targeted non-reasoning tasks, for example, language modeling, open-ended generation, and summarization. These tasks are less sensitive to distributional sparsity and often benefit from simple accumulation or semantic filtering strategies that enhance diversity.

However, our work demonstrates that reasoning tasks are more vulnerable to model collapse due to complex reasoning tasks require structured, compositional, and often multi-hop inference capabilities. In such tasks, simply expanding diversity in a linguistic or syntactic manner often filters out complete and structured reasoning trajectories or repeat low quality data, with models failing to learn high-quality reasoning patterns with generalization. This is a critical gap that previous literature has not examined, especially in SFT reasoning data.
This work diagnoses and mitigates model collapse for reasoning tasks under SFT, showing that naive distribution expansion can be ineffective or even worse. Our model interpretability analysis shows that using self-generated data will strengthen the harmful model weights updating. Therefore, data diversity method still hard to avoid it. In contrast, we propose that model merging provides a more robust and effective solution inspired by interpretability analysis. Empirically, we show that this strategy outperforms data-centric diversity enhancements in preserving general reasoning capability. This distinction between task types (reasoning vs. non-reasoning), model interpretability, and the limitations of prior methods is an important motivation of our work.

\subsection{Related Works on LLM Reasoning}
\label{app:realted_works}
LLMs have demonstrated remarkable success across various reasoning tasks, including mathematical problem-solving, code generation, multi-modality, agent, and common-sense reasoning \cite{yu2024metamath, wang2023codet5+, shi2023crossget, diao2025soundmind, fu2024amoeballm, wu2025assessing, shilacache, jin2025two, ye2025longmamba, diao2025soundmind, zhang2025overcoming, zhang2025growing}. Beyond leveraging sophisticated prompting techniques to enhance reasoning capabilities \cite{kojima2022large, wei2022chain, zheng2023take, yao2024tree}, many methods focus on fine-tuning LLMs with reasoning datasets to create more robust reasoners \cite{lu2024mathgenie, yu2024metamath}. For instance, approaches like SI \cite{huang2022large}, STaR \cite{zelikman2022star}, V-STaR \cite{hosseini2024v}, and rSTaR \cite{qi2025mutual} fine-tune LLMs on task-specific datasets or synthesize reasoning data tailored for corresponding tasks.
In addition to training models to generate correct answers, some studies introduce external verifiers \cite{cobbe2021training, lightman2023let, hosseini2024v, yuan2024self} that select the best answer from a set of candidate solutions.

\section{Potential Risks}
Enhancing LLMs with self-improving generalized reasoning capability may introduce risks of unintended capability emergence, including misuse in adversarial contexts such as misinformation or manipulation. As the model gains broader reasoning abilities across domains, it may be used for enabling harmful applications with enhanced reasoning capability. This highlights the importance of pairing IMM with safe evaluation and alignment to ensure safe and responsible deployment.

\begin{table*}[]
\resizebox{\textwidth}{!}{
\begin{tabular}{l}
\toprule
\textbf{Prompt for Generating GSM8K Answers}

\\ \midrule
\begin{tabular}[c]{@{}l@{}}Below is an instruction that describes a task.  \\ Write a response that appropriately completes the request like given examples below:\\ \\ Question: Angelo and Melanie want to plan how many hours over the next week they should study together for their test next week. \\ They have 2 chapters of their textbook to study and 4 worksheets to memorize. They figure out that they should dedicate 3 hours to \\ each chapter of their textbook and 1.5 hours for each worksheet. If they plan to study no more than 4 hours each day, how many days \\ should they plan to study total over the next week if they take a 10-minute break every hour, include 3 10-minute snack breaks each \\ day, and 30 minutes for lunch each day?\\ \\ A: Let's think step by step.\\ Angelo and Melanie think they should dedicate 3 hours to each of the 2 chapters, 3 hours x 2 chapters = 6 hours total.\\ For the worksheets they plan to dedicate 1.5 hours for each worksheet, 1.5 hours x 4 worksheets = 6 hours total.\\ Angelo and Melanie need to start with planning 12 hours to study, at 4 hours a day, 12 / 4 = 3 days.\\ However, they need to include time for breaks and lunch. Every hour they want to include a 10-minute break, so 12 total hours x 10 \\ minutes = 120 extra minutes for breaks.\\ They also want to include 3 10-minute snack breaks, 3 x 10 minutes = 30 minutes.\\ And they want to include 30 minutes for lunch each day, so 120 minutes for breaks + 30 minutes for snack breaks + 30 minutes for l\\ unch = 180 minutes, or 180 / 60 minutes per hour = 3 extra hours.\\ So Angelo and Melanie want to plan 12 hours to study + 3 hours of breaks = 15 hours total.\\ They want to study no more than 4 hours each day, 15 hours / 4 hours each day = 3.75\\ They will need to plan to study 4 days to allow for all the time they need.\\ The answer is 4\\ \\ Question: Mark's basketball team scores 25 2 pointers, 8 3 pointers and 10 free throws.  Their opponents score double the 2 pointers \\ but half the 3 pointers and free throws.  What's the total number of points scored by both teams added together?\\ A: Let's think step by step.\\ Mark's team scores 25 2 pointers, meaning they scored 25*2= 50 points in 2 pointers.\\ His team also scores 6 3 pointers, meaning they scored 8*3= 24 points in 3 pointers\\ They scored 10 free throws, and free throws count as one point so they scored 10*1=10 points in free throws.\\ All together his team scored 50+24+10= 84 points\\ Mark's opponents scored double his team's number of 2 pointers, meaning they scored 50*2=100 points in 2 pointers.\\ His opponents scored half his team's number of 3 pointers, meaning they scored 24/2= 12 points in 3 pointers.\\ They also scored half Mark's team's points in free throws, meaning they scored 10/2=5 points in free throws.\\ All together Mark's opponents scored 100+12+5=117 points\\ The total score for the game is both team's scores added together, so it is 84+117=201 points\\ The answer is 201\\ \\ Question: Bella has two times as many marbles as frisbees. She also has 20 more frisbees than deck cards. If she buys 2/5 times \\ more of each item, what would be the total number of the items she will have if she currently has 60 marbles?\\ A: Let's think step by step.\\ When Bella buys 2/5 times more marbles, she'll have increased the number of marbles by 2/5*60 = 24\\ The total number of marbles she'll have is 60+24 = 84\\ If Bella currently has 60 marbles, and she has two times as many marbles as frisbees, she has 60/2 = 30 frisbees.\\ If Bella buys 2/5 times more frisbees, she'll have 2/5*30 = 12 more frisbees.\\ The total number of frisbees she'll have will increase to 30+12 = 42\\ Bella also has 20 more frisbees than deck cards, meaning she has 30-20 = 10 deck cards\\ If she buys 2/5 times more deck cards, she'll have 2/5*10 = 4 more deck cards.\\ The total number of deck cards she'll have is 10+4 = 14\\ Together, Bella will have a total of 14+42+84 = 140 items\\  The answer is 140\\ \\ Other 5 examples here ...\\ \\ \#\#\# Instruction:\\ Natalia sold clips to 48 of her friends in April, and then she sold half as many clips in May. How many clips did Natalia sell \\ altogether in April and May?\\ \\ \#\#\# Response: Let's think step by step.\end{tabular} \\ \bottomrule
\end{tabular}}
\caption{The CoT prompting examples for generate training data.}
\label{tab:cot_gsm8k}
\end{table*}

\end{document}